\documentclass[conference]{IEEEtran}
\ifCLASSINFOpdf
   \usepackage[pdftex]{graphicx}
   \usepackage{multirow}
\else
\fi
\usepackage[tight,footnotesize]{subfigure}

\usepackage{color}
\usepackage{multirow}

\usepackage{enumitem}
\usepackage{flushend}
\AtEndDocument{\par\leavevmode}

\IEEEoverridecommandlockouts

\begin{document}
%
\title{Evaluation of Pose Tracking Accuracy in the\\First and Second Generations of Microsoft Kinect}

\author{\IEEEauthorblockN{Qifei Wang \& Gregorij Kurillo}
\IEEEauthorblockA{University of California, Berkeley\\
	Department of EECS\\
	Berkeley, CA 94720, USA\\
Email: \{qifei.wang, gregorij\}@eecs.berkeley.edu}
\and
\IEEEauthorblockN{Ferda Ofli}
\IEEEauthorblockA{Qatar Computing Research Institute\\
	Hamad Bin Khalifa University\\
	Doha, Qatar\\
Email: fofli@qf.org.qa}
\and
\IEEEauthorblockN{Ruzena Bajcsy}
\IEEEauthorblockA{University of California, Berkeley\\
	Department of EECS\\
	Berkeley, CA 94720, USA\\
	Email: bajcsy@eecs.berkeley.edu}
\thanks{This research was supported by the National Science Foundation (NSF) under Grant No. 1111965.}}


%


\def\etal{\emph{et al.}}

\newcommand*{\comment}{\textcolor{red}}

\maketitle

\begin{abstract}
Microsoft Kinect camera and its skeletal tracking capabilities have been embraced by many researchers and commercial developers in various applications of real-time human movement analysis. In this paper, we evaluate the accuracy of the human kinematic motion data in the first and second generation of the Kinect system, and compare the results with an optical motion capture system. 
We collected motion data in 12 exercises for 10 different subjects and from three different viewpoints. We report on the accuracy of the joint localization and bone length estimation of Kinect skeletons in comparison to the motion capture. 
We also analyze the distribution of the joint localization offsets by fitting a mixture of Gaussian and uniform distribution models to determine the outliers in the Kinect motion data. 
Our analysis shows that overall Kinect 2 has more robust and more accurate tracking of human pose as compared to Kinect 1.
\end{abstract}


%
\IEEEpeerreviewmaketitle

	\section{Introduction}
	\label{sec:intro}

Affordable markerless motion capture technology is becoming increasingly pervasive in applications of human-computer and human-machine interaction, entertainment, healthcare, communication, surveillance and others. Although the methods for capturing and extracting human pose from image data have been around for several years, the advances in sensor technologies (infrared sensors) and computing power (e.g., GPUs) have facilitated new systems that provide robust and relatively accurate markerless acquisition of human movement. An important milestone for wide adoption of these technologies was the release of Microsoft Kinect camera~\cite{Zhang_2012} for the gaming console Xbox 360 in 2010, followed by the release of Kinect for Windows with the accompanying Software Development Kit (SDK) in 2011. The Kinect SDK for Windows featured real-time full-body tracking of human limbs based on the algorithm by Shotton \etal~\cite{Shotton_2011}. Several other technology makers followed the suit by releasing their own 3D cameras that focused on capture of human motion for interactive applications (Xtion by Asus, RealSense by Intel). Many researchers and commercial developers embraced the Kinect in wide range of applications that took advantage of its real-time 3D acquisition capabilities and provided skeletal tracking, such as in physical therapy and rehabilitation~\cite{Hondori2014}, fall detection~\cite{Stone2011} and exercise in elderly~\cite{Webster2014,Ofli2015}, ergonomics~\cite{Diego-Mas2013, Plantard2015} and anthropometry~\cite{Robinson2013}, computer vision~\cite{Han2013}, and many others. In 2013 the second generation of the Kinect camera was released as part of the Xbox One gaming console. In 2014 a standalone version of Kinect for Windows (k4w) was officially released featuring wider camera angle, higher resolution of depth and color images, improved skeletal tracking, and detection of facial expressions.

In this paper we focus on the evaluation of accuracy and performance of skeletal tracking in the two Kinect systems (referred to as \emph{Kinect 1} and \emph{Kinect 2} in the remainder of this paper) compared to a marker-based motion capture system. Several publications have previously addressed the accuracy of the skeletal tracking of Kinect 1 for various applications; however, the accuracy of Kinect 2 has been reported only to a limited extent in the research literature. Furthermore, concurrent comparison of the two systems has not yet been published to the best of our knowledge. Although both Kinect systems employ similar methodology for human body segmentation and tracking based on the depth data, the underlying technology for acquisition of the depth differs between the two. 
We report the accuracy rates of the skeletal tracking and the corresponding error distributions in a set of exercise motions that include standing and sitting body configurations. Such an extensive performance assessment of the technology is intended to assist researchers who rely on Kinect as a measurement device in the studies of human motion.

	\section{Related Work}
	\label{sec:related_work}

	In this section we review several publications related to evaluation of the Kinect systems. Kinect 1 has been extensively investigated in terms of 3D depth map acquisition as well as body tracking accuracy for various applications. Khoshelman and Elbernik~\cite{Khoshelham_2013} examined the accuracy of depth acquisition in Kinect 1, and found that the depth error ranges from a few millimeters up to about 4 cm at the maximum range. They recommended that the data for mapping applications should be acquired within 1-3 m distance. Smisek \etal~\cite{Smisek_2011} proposed a geometrical model and calibration method to improve the accuracy of Kinect 1 for 3D measurements. Kinect 1 and Kinect 2 were jointly evaluated by Gonzalez-Jorge \etal~\cite{Gonzalez_2015} who reported that the precision of both systems is similar (about 8 mm) in the range of under 1 m, while Kinect 2 outperforms Kinect 1 at the range of 2 m with the error values of up to 25 mm. They also reported that precision of Kinect 1 decreases rapidly following a second order polynomial, while Kinect 2 exhibits a more stable behavior inside its work range. 3D accuracy of Kinect 2 was recently evaluated by Yang \etal~\cite{Yang_2015} who reported on the spatial distribution of the depth accuracy in regard to the vertical and horizontal displacement.
	
	Skeletal tracking of Kinect was examined primarily in the context of biomechanical and exercise performance analyses. In this review, we limit ourselves only to the evaluations of skeletal tracking based on the official Microsoft Kinect SDK. Obdr\v{z}\'{a}lek \etal~\cite{Obdrzalek_2012} performed accuracy and robustness analysis of the Kinect skeletal tracking in six exercises for elderly population. Their paper reports on the error bounds for particular joints obtained from the comparison with an optical motion capture system. The authors conclude that employing a more anthropometric kinematic model with fixed limb lengths could improve the performance.
	Clark \etal~\cite{Clark_2012} examined the clinical feasibility of using Kinect for postural control assessment. The evaluation with Kinect and motion capture performed in 20 healthy subjects included three postural tests: forward reach, lateral reach, and single-leg eyes-closed standing balance assessment. The authors found high inter-trail reliability and excellent concurrent validity for majority of the measurements. The study, however, revealed presence of proportional biases for some of the outcome measures, in particular for sternum and pelvis evaluations. The authors proposed the use of calibration equations that could potentially correct for such biases.
	Several other works have examined the body tracking accuracy for specific applications in physical therapy, such as for example upper extremity function evaluation~\cite{Kurillo_2013}, assessment of balance disorders~\cite{Funaya_2013}, full-body functional assessment~\cite{Bonnechere_2013}, and movement analysis in Parkinson's disease~\cite{Galna_2014}. Plantard \etal~\cite{Plantard_2015} performed an extensive evaluation of Kinect 1 skeletal tracking accuracy for ergonomic assessment. By using a virtual mannequin, they generated a synthetic depth map that was input into the Kinect SDK algorithm to predict potential accuracy of joint locations in a large number of skeletal configurations and camera positions. The simulation results were validated by a small number of real experiments. The authors concluded that the kinematic information obtained by the Kinect is generally accurate enough for ergonomic assessment.
	
	To the best of our knowledge, publication by Xu and McGorry~\cite{Xu_2015} is to date the only work that reported on the evaluation of Kinect 2 skeletal tracking alongside an optical motion capture system. In their study the authors examined 8 standing and 8 sitting static poses of daily activities. Similar poses were also captured with Kinect 1, however the two data sets were not obtained concurrently. The authors reported that the average static error across all the participants and all Kinect-identified joint centers was 76 mm for Kinect 1 and 87 mm for Kinect 2. They further concluded that there was no significant difference between the two Kinects. This conclusion, however, is of limited validity as the comparison was done indirectly with two different sets of subjects.
	
	Since the Kinect~1 system is being replaced by the Kinect~2 system in many applications, it is important to evaluate the performance of the new camera and software for tracking of dynamic human activities. This is especially relevant since the depth estimation in the two systems is based on two different physical principles. Side-by-side comparison can thus provide a better understanding of the performance improvements as well as potential drawbacks. In this paper, we report on the experimental evaluation of the joint tracking accuracy of the two Kinects in comparison to an optical motion capture system. We analyze the results for 12 different activities that include standing and sitting poses as well as slow and fast movements. Furthermore we examine the performance of pose estimation with respect to three different horizontal orientation angles. We provide the error bounds for joint positions and extracted limb lengths for both systems. We also analyze the distribution of joint localization errors by fitting a mixture of Gaussian and uniform distribution models to determine the outliers in the motion data.

	\section{Acquisition Systems}
	\label{sec:systems}
	In this section we provide more details on the experimental setup and a brief overview of the technology behind each Kinect system. For the experimental evaluation, the movements were simultaneously captured by Kinect~1, Kinect~2, and a marker-based optical motion capture system which served as a baseline. The two Kinects were secured together and mounted on a tripod at the height of about 1.5 m.  All three systems were geometrically calibrated and synchronized prior to the data collection using the procedure described below. 
	
	\subsection{Motion Capture System (MoCap)}
	The motion capture data were acquired using PhaseSpace (San Leandro, CA, USA) system Impulse X2 with 8 infrared stereo cameras. The cameras were positioned around the capture space of about 4 m by 4 m. The system provides 3D position of LED markers with sub-millimeter accuracy and frequency of up to 960 Hz. Capture rate of 480 Hz was selected for this study. For each subject 43 markers were attached at standard body landmarks to a motion capture suit using velcro. To obtain the skeleton from the marker data, a rigid kinematic structure was dynamically fitted into the 3D point cloud. We used PhaseSpace Recap2 software to obtain the skeleton for each subject based on collected calibration data which consisted of a sequence of individual joint rotations. The built-in algorithm determines the length of the body segments based on the set of markers associated with different parts of the body and generates a skeleton with 29 joint positions. Once individual's kinematic model is calibrated, the skeletal sequence can be extracted for any motion of that person.
	
	\subsection{Kinect~1}
	Kinect~1 sensor features acquisition rates of up to 30 Hz for the color and depth data with the resolution of 640 $\times$ 480 pixels and 320 $\times$ 240 pixels, respectively. The depth data are obtained using structured light approach, where a pseudo-random infrared dot pattern is projected onto the scene while being captured by an infrared camera. Stereo triangulation is used to obtain 3D position of the points from their projections. This approach provides a robust 3D reconstruction even in low-light conditions. The accuracy of the depth decreases with the square of the distance with typical accuracy ranging from about 1-4 cm in the range of 1-4 m~\cite{Smisek_2011}. To obtain a dense depth map, surface interpolation is applied based on the acquired depth values at the data points. Fixed density of the points limits the accuracy when moving away from the camera as the points become sparser. The boundaries of surfaces in the distance are thus often jagged. 
	
	Real-time skeletal tracking provided by the Microsoft Kinect SDK is based on the depth data using body part estimation algorithm based on random decision forest proposed by Shotton \etal~\cite{Shotton_2011}. The algorithm estimates candidate body parts based on a large training set of synthetically-generated depth images of humans of many different poses and shapes in various poses from a motion capture database~\cite{Zhang_2012}. The Kinect~1 SDK can track up to two users, providing the 3D location of 20 joints for each tracked skeleton.

	\subsection{Kinect~2}
	Kinect~2 sensor features high definition color (1920 $\times$ 1080 pixels) and higher resolution depth data (512 $\times$ 424 pixels) as compared to Kinect~1. The depth acquisition is based on the time-of-flight (ToF) principle where the distance to points on the surface is measured by computing the phase-shift distance of modulated infrared light. The intensity of the captured image is thus proportional to the distance of the points in 3D space. The ToF technology as opposed to the structured light inherently provides a dense depth map, however the results can suffer from various artifacts caused by the reflections of light signal from the scene geometry and the reflectance properties of observed materials. The depth accuracy of Kinect~2 is relatively constant within a specific capture volume, however it depends on the vertical and horizontal displacement as the light pulses are scattered away from the center of the camera~\cite{Yang_2015}. The reported average depth accuracy is under 2 mm in the central viewing cone and increases to 2-4 mm in the range of up to 3.5 m. The maximal range captured by Kinect~2 is 4.0 m where the average error typically increases beyond 4 mm.
	
	The skeletal tracking method implemented in Kinect SDK v2.0 has not been fully disclosed by Microsoft; however, it appears to follow similar methodology as for Kinect~1 while taking advantage of GPU computation to reduce the latency and to improve the performance. The Kinect SDK v2.0 features skeletal tracking of up to 6 users with 3D locations of 25 joints for each skeleton. In comparison to Kinect~1, the skeleton includes additional joints at the hand tip, thumb tip and neck. The arrangement of the joints, i.e. the kinematic structure of the model, is comparable to standard motion capture skeleton. Kinect~2 includes some additional features, such as detection of hand opening/closing and tracking of facial features.

	\subsection{Calibration and Data Acquisition}
	\label{ssec:Calibration}
	
	For the capture of the database, we connected the two Kinect cameras to a single PC running Windows 8.1, with Kinect~1 connected via USB 2.0 and Kinect~2 connected via USB 3.0 on a separate PCI bus. Such arrangement allowed for both sensors to capture at the full frame rate of 30 Hz. The skeletal data for both cameras were extracted in real time via Microsoft Kinect SDK v1.8 and Kinect for Windows SDK v2.0 for Kinect~1 and Kinect~2, respectively. 
	
	The temporal synchronization of the captured data was performed using Network Time Protocol (NTP). The motion capture server provided the time stamps for the Kinect PC over the local area network. Meinberg NTP Client Software (Meinberg Radio Clocks GmbH, Bad Pyrmont, Germany) was installed on the Windows computer to obtain more precise clock synchronization.
	
	Prior to the data acquisition, we first calibrated the motion capture system using provided calibration software. The coordinate frames of the Kinect cameras were then aligned with the motion capture coordinates using the following procedure. A planar checkerboard with three motion capture markers attached to corners of the board was placed in three different positions in front of the Kinects. In each configuration, marker position, color and depth data were recorded. Next, the 3D positions of the corners were extracted from the depth data using the intrinsic parameters of the Kinect and corresponding depth pixel values. Finally, a rigid transformation matrix that maps 3D data captured by each Kinect into the motion capture coordinate system was determined by minimizing the squared distance between the Kinect acquired points and the corresponding marker locations.

	\begin{figure}[!htbp]
		\centering 
		\includegraphics[trim=120 40 70 40, clip, width=6cm] {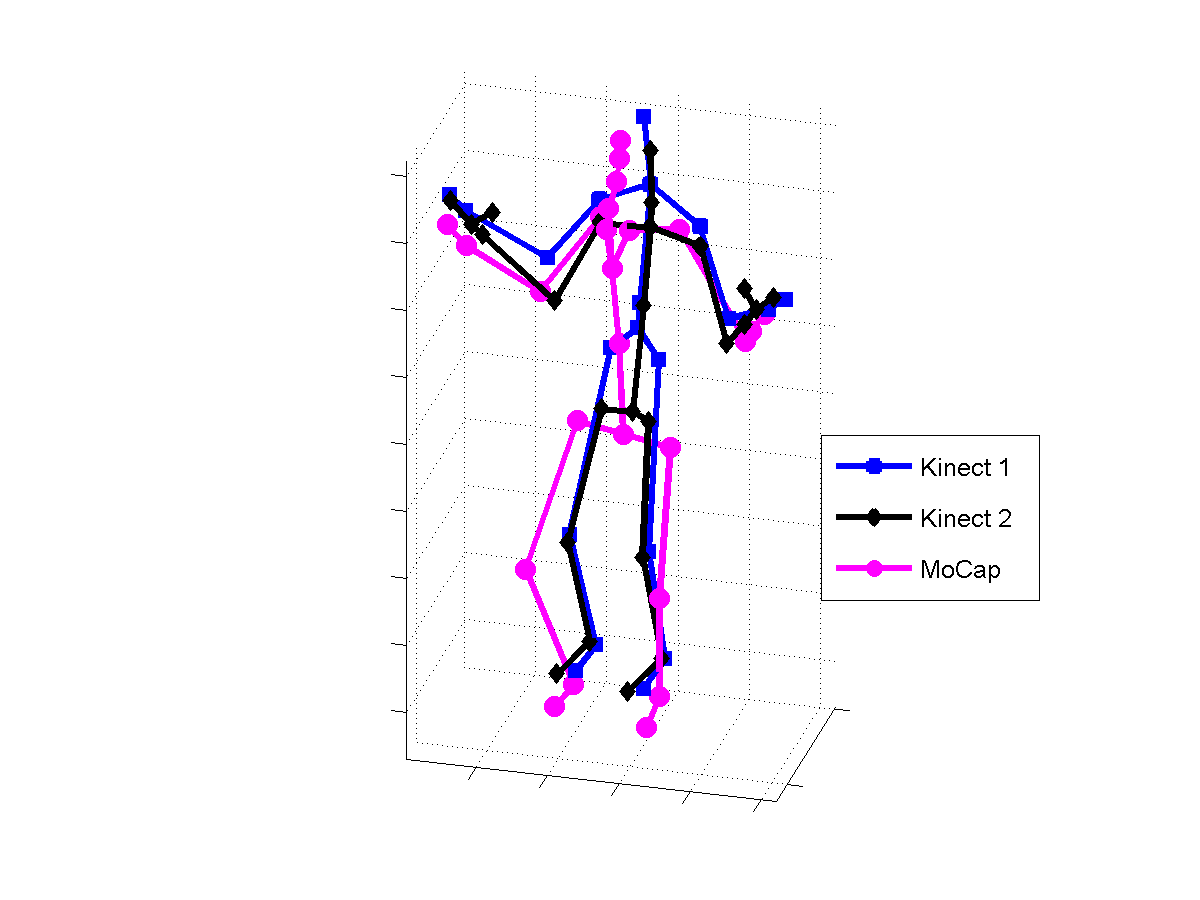}
		\caption{Three skeletons captured by Kinect~1, Kinect~2, and motion capture (extracted via Recap2 software) after geometric and temporal alignment.}
		\label{fig:skeleton_demo}
	\end{figure}

	\subsection{Data Processing}
	
	Collected marker data were first processed in Recap2 to obtain the skeletal sequence for each subject, and then exported to BVH file format. The rest of the analysis was performed in MATLAB (MathWorks, Natick, MA). First, the skeletal sequences from the Kinect cameras were mapped to the motion capture coordinate space using the rigid transformation obtained from the calibration. Next, we aligned the sequences using the time stamps, and re-sampled all the data points to the time stamps of Kinect~2 in order to compare the joint localization at the same time instances. Fig.~\ref{fig:skeleton_demo} demonstrates the three skeletal configurations projected into the motion capture coordinate space after the calibration. 
	
	After the spatial transformation and temporal alignment, we obtained three sequences of 3D joint positions for Kinect~1, Kinect~2, and motion capture. Since the three skeletal configurations have different arrangements and number of joints, we selected 20 joints that are common to all the three systems. Other remaining joints were ignored in this analysis. Next, we evaluated the position accuracy by calculating the distance between the corresponding joints in each time frame. When the Kinect skeletal tracking loses track of the body parts for certain joints (e.g. due to occlusions), such frames can be flagged as outliers. Since the data of the outliers can be assigned arbitrary values, we use a uniform distribution to model the distribution of the outliers. The distribution of the valid (on-track) data samples is on the other hand modeled by a Gaussian distribution with the mean representing the average offset of that joint. The overall distribution of the joint offset data, $p(\theta)$, can thus be modeled by a mixture model of Gaussian and uniform distributions as follows:
	\begin{equation}
		p(\theta) = \rho\times N(\mu,\sigma)+(1-\rho)\times U(x_1, x_2).
		\label{equ:mixture_model}
	\end{equation}
	In equation~(\ref{equ:mixture_model}), $\mu$ and $\sigma$ denote the parameters of the Gaussian distribution, $N$, respectively. $x_1$ and $x_2$ denote the parameters of the uniform distribution, $U$, respectively. $\rho$ denotes the weight of the Gaussian distribution. In this paper, we use the maximum-likelihood method to estimate these parameters with the input data samples. After estimating the mixture model, the data are classified into either \textit{on-track} or \textit{off-track} state. The off-track data is then excluded from the accuracy evaluation.
	
	Another important parameter for the accuracy assessment of human pose tracking is the variability of the limb lengths. The human skeleton is typically modeled as a kinematic chain with rigid body segments. The Kinect skeletal tracking, however, does not explicitly constrain the length of body segments. In the paper, we thus report on the variability of the bone lengths by calculating the distance between two end-joints of each bone for the Kinect systems. For motion capture, the bone length is extracted from the segment length parameters in the BVH file.

	\section{Experiments}
	\label{sec:experiments}
\begin{figure*}[t]
	\centering     
	\subfigure[Shallow Squats]{\label{fig:a}
		\includegraphics[trim=0 0 0 100, clip, width=0.15\textwidth]{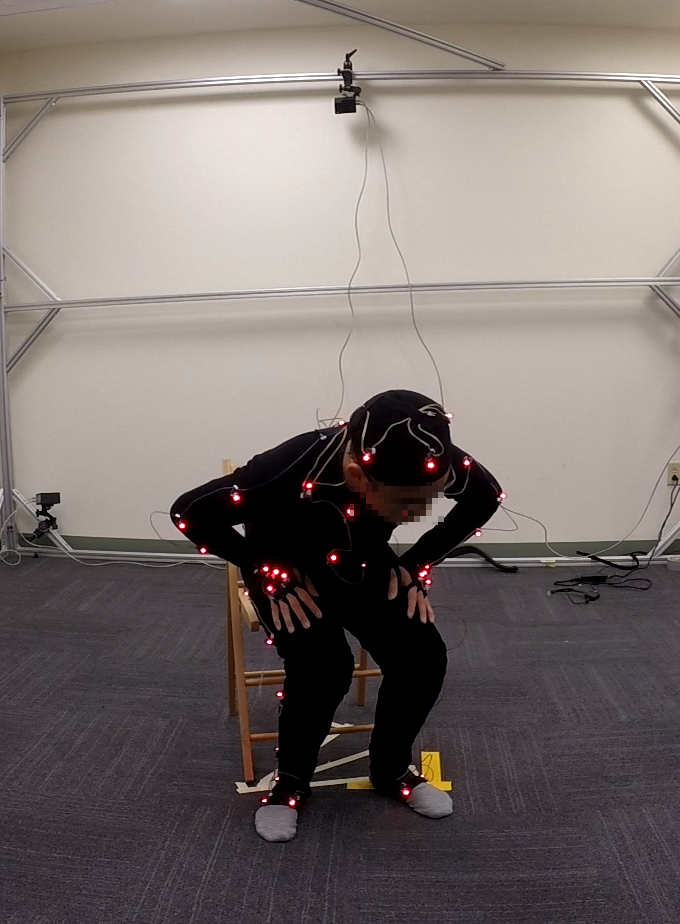}}
	\subfigure[Chair Stands ]{\label{fig:b}
		\includegraphics[trim=0 0 0 100, clip, width=0.15\textwidth]{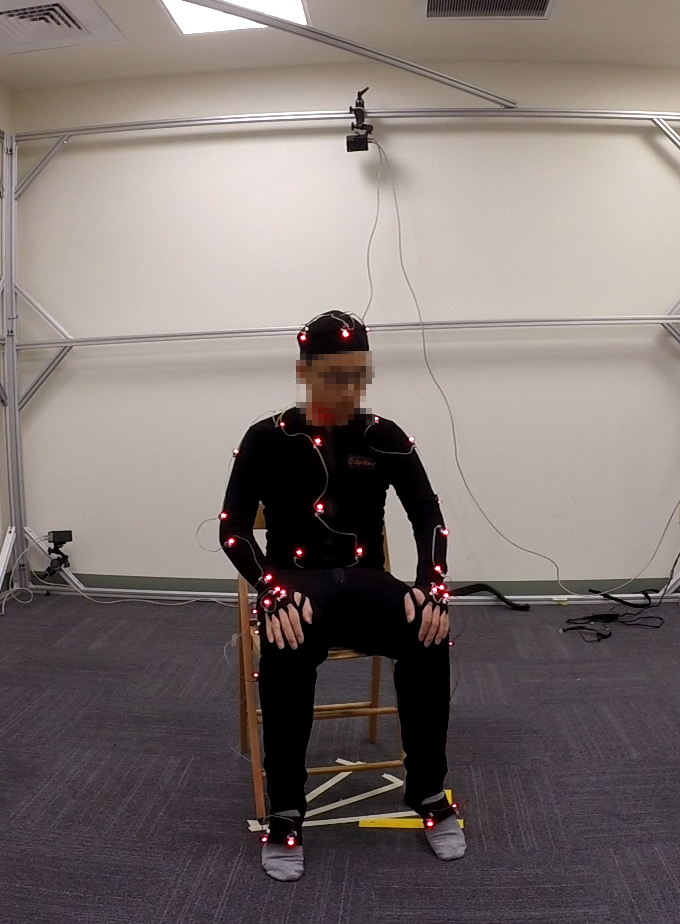}}
	\subfigure[Buddha's Prayer]{\label{fig:c}
		\includegraphics[trim=0 0 0 100, clip, width=0.15\textwidth]{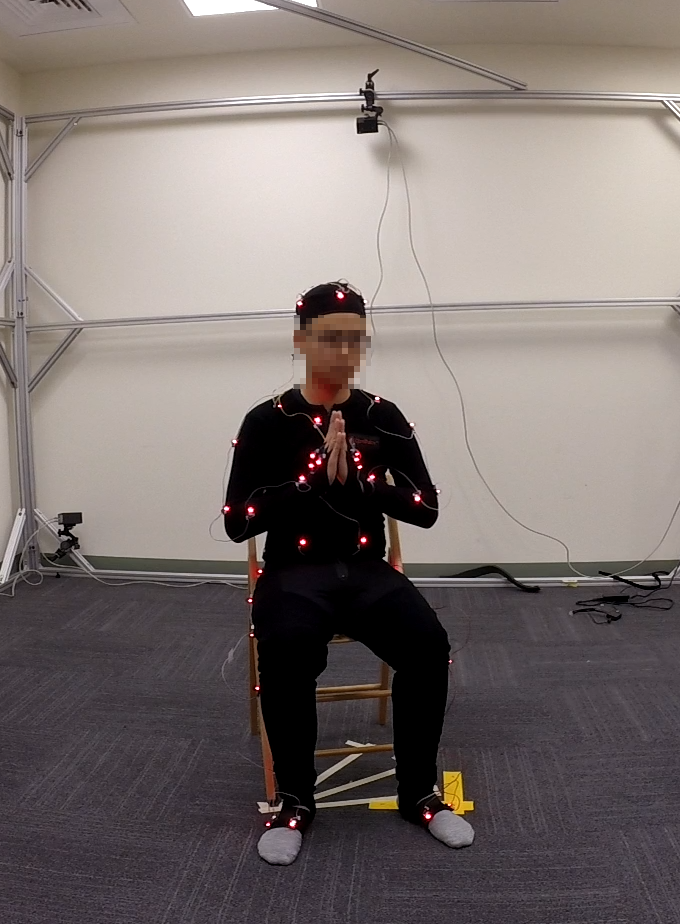}}
	\subfigure[Cops \& Robbers]{\label{fig:d}
		\includegraphics[trim=0 0 0 100, clip, width=0.15\textwidth]{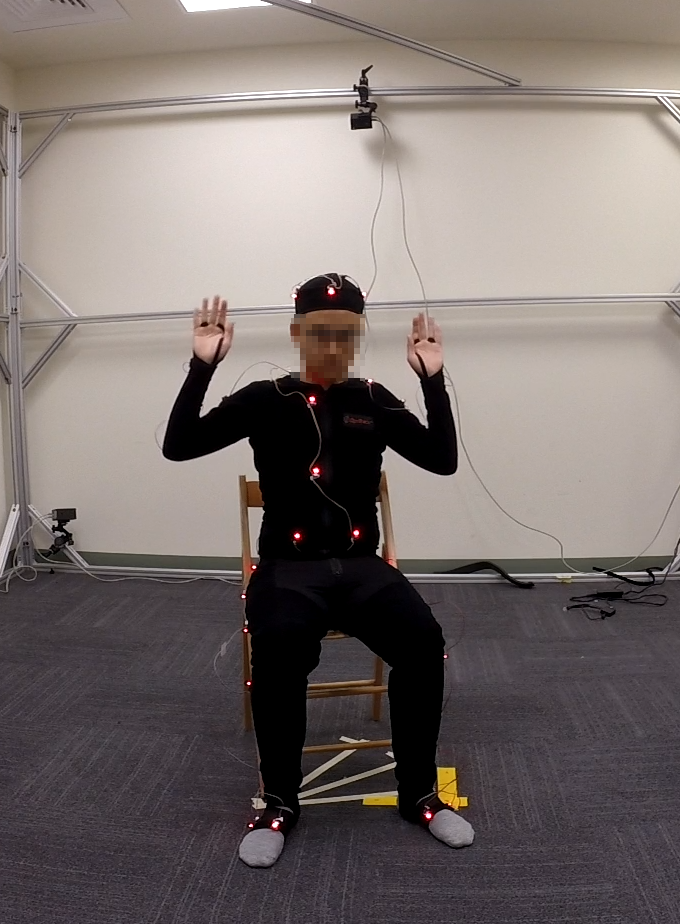}}
	\subfigure[Abs in, Knee Lifts]{\label{fig:e}
		\includegraphics[trim=0 0 0 100, clip, width=0.15\textwidth]{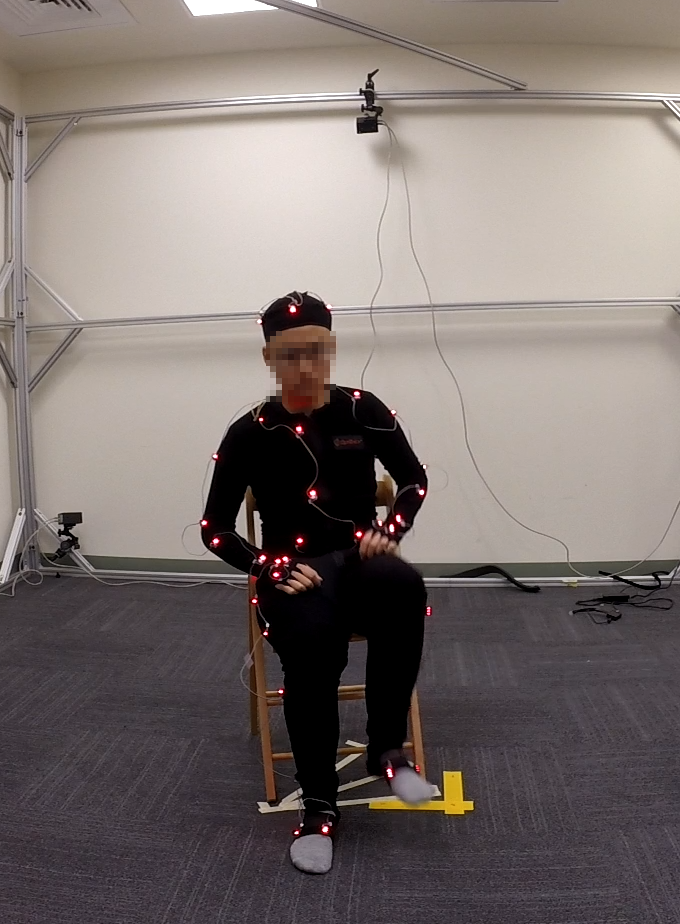}}
	\subfigure[Lateral Stepping]{\label{fig:f}
		\includegraphics[trim=0 0 0 100, clip, width=0.15\textwidth]{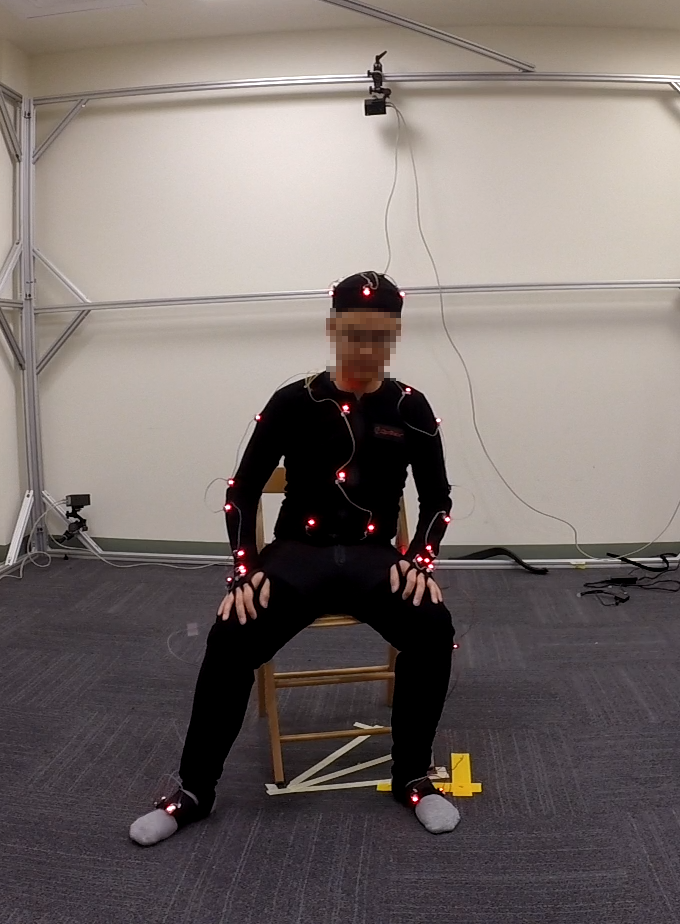}}
	\\
	\subfigure[Pickup \& Throw]{\label{fig:g}
		\includegraphics[trim=0 0 0 100, clip, width=0.15\textwidth]{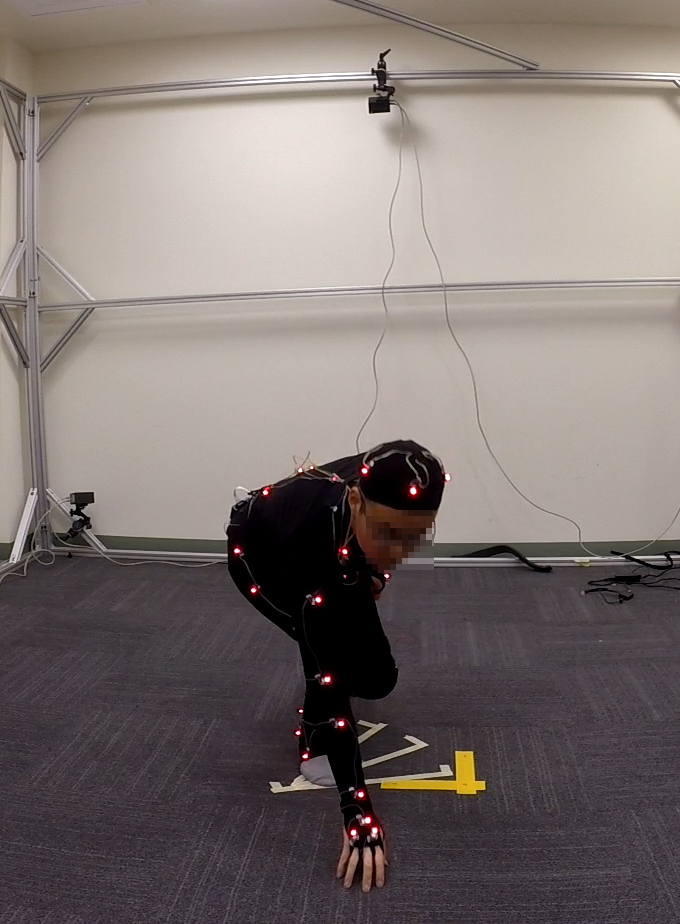}}
	\subfigure[Jogging]{\label{fig:h}
		\includegraphics[trim=0 0 0 100, clip, width=0.15\textwidth]{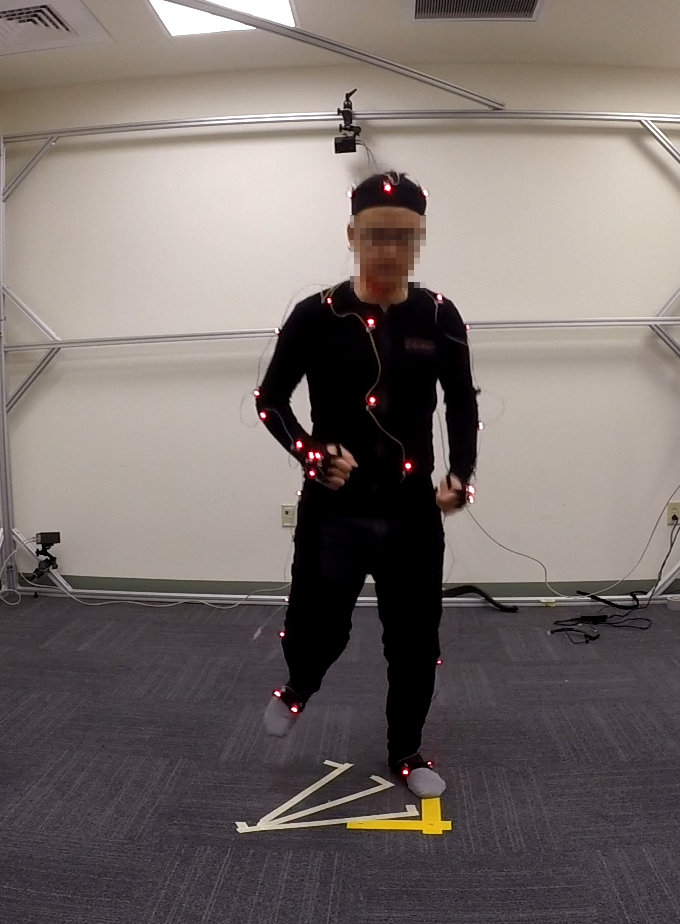}}
	\subfigure[Clapping]{\label{fig:i}
		\includegraphics[trim=0 0 0 100, clip, width=0.15\textwidth]{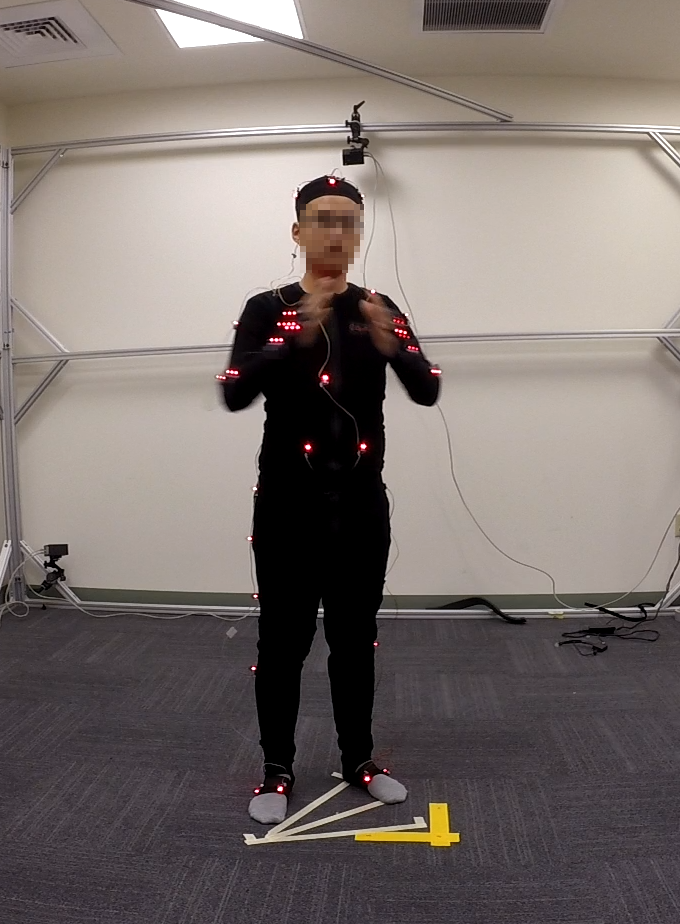}}
	\subfigure[Punching]{\label{fig:j}
		\includegraphics[trim=0 0 0 100, clip, width=0.15\textwidth]{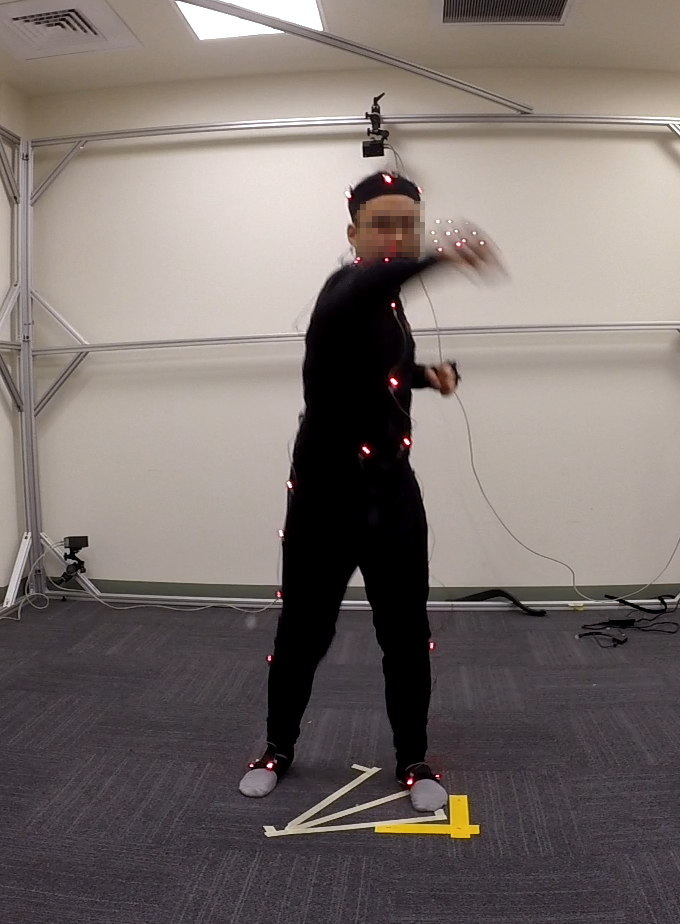}}
	\subfigure[Line Stepping]{\label{fig:k}
		\includegraphics[trim=0 0 0 100, clip, width=0.15\textwidth]{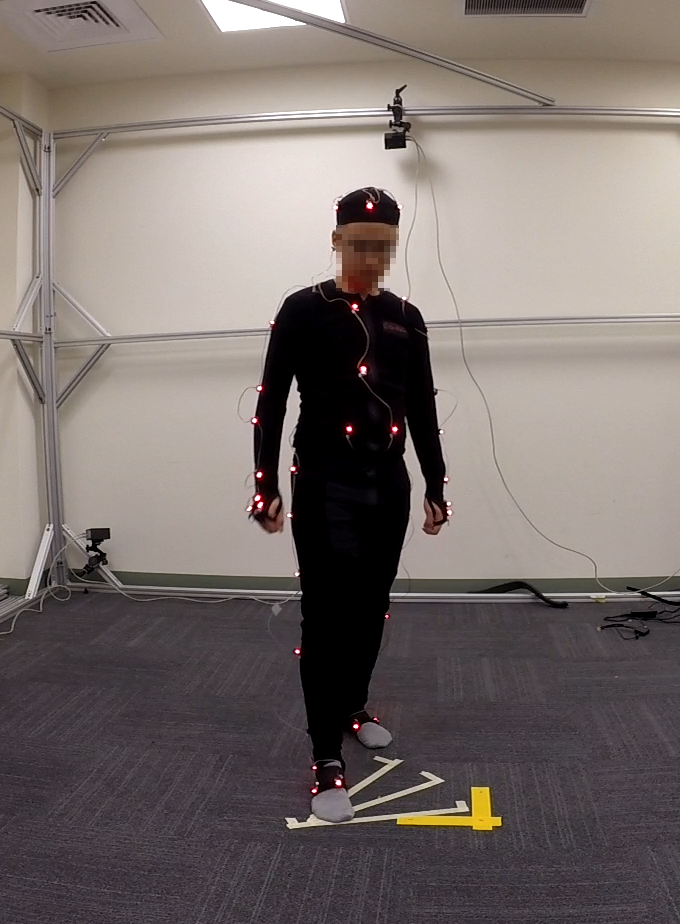}}
	\subfigure[Pendulum]{\label{fig:l}
		\includegraphics[trim=0 0 0 100, clip, width=0.15\textwidth]{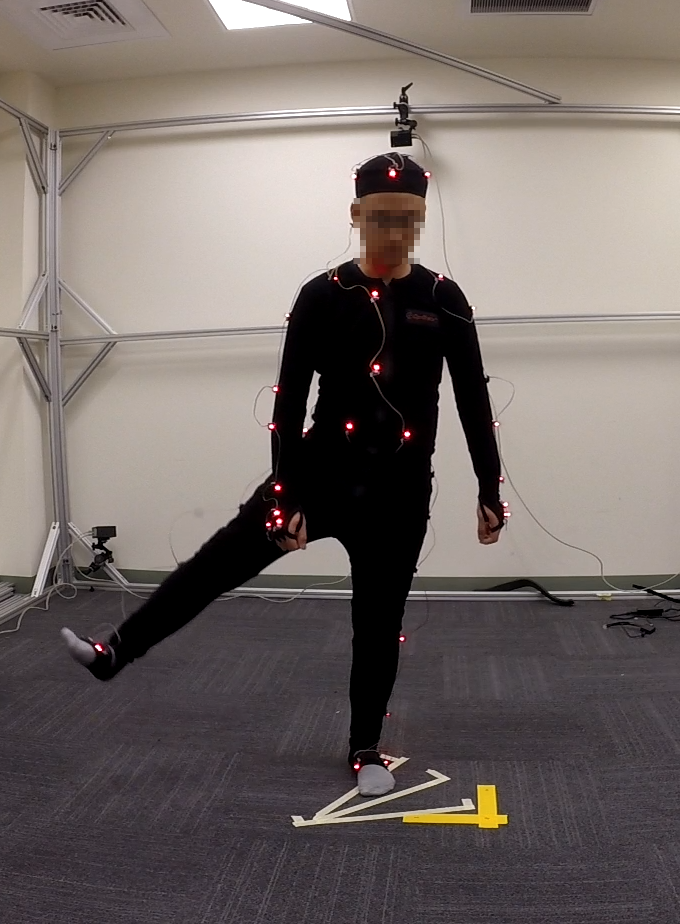}}
	\caption{Video snapshots of the 12 exercises. The first set (a-f) consisted of seated exercises, while the second set (g-l) consisted of standing exercises.}
	\label{fig:exer_snapshots}
\end{figure*}

In this section we describe the experimental protocol for the data accuracy evaluation. As described in Section \ref{sec:systems}, the motion data were captured by the setup consisting of Kinect~1, Kinect~2, and the motion capture system. We selected 12 different exercises (Table~\ref{table:exercise_list}, Fig.~\ref{fig:exer_snapshots}), consisting of six sitting (and sit-to-stand) exercises and six standing exercises. In the first set of exercises, subjects were interacting with the chair, while no props were used in the second set. We analyze the two sets of exercises separately. 

\begin{table}[!htbp] 
	\centering
	\caption{List of exercises.}
	\label{table:exercise_list}
	\begin{tabular}{|l|l|l|}
		\hline
		Name & Pose & Description \\
		\hline\hline
		1. Shallow Squats & Sitting & Stand-to-sit movements without sitting.\\
		2. Chair Stands   & Sitting  & Sit-to-stand movements.\\
		3. Buddha's Prayer      & Sitting  & Vertical hand lifts with palms together.\\
		4. Cops \& Robbers     & Sitting & Shoulder rotation and forward arm extension. \\
		5. Abs in, Knee Lifts    & Sitting  & Alternating knee lifts.\\
		6. Lateral Stepping     & Sitting  & Alternating front and side stepping.\\
		7. Pickup \& Throw   & Standing & Step forward, pick up off the floor and throw. \\
		8. Jogging            & Standing  & Running in place. \\
		9. Clapping           & Standing  & Wide hand clapping while standing still. \\
		10. Punching          & Standing  & Alternating forward punching. \\
		11. Line Stepping     & Standing  & Alternating forward foot tapping.\\
		12. Pendulum      & Standing   & Alternating leg adduction.\\
		\hline
	\end{tabular}
\end{table}

\begin{figure*}[t]
	\centering     
	\subfigure[Shallow Squats]{\label{fig:a}
		\includegraphics[trim=0 30 0 100, clip, width=0.15\textwidth]{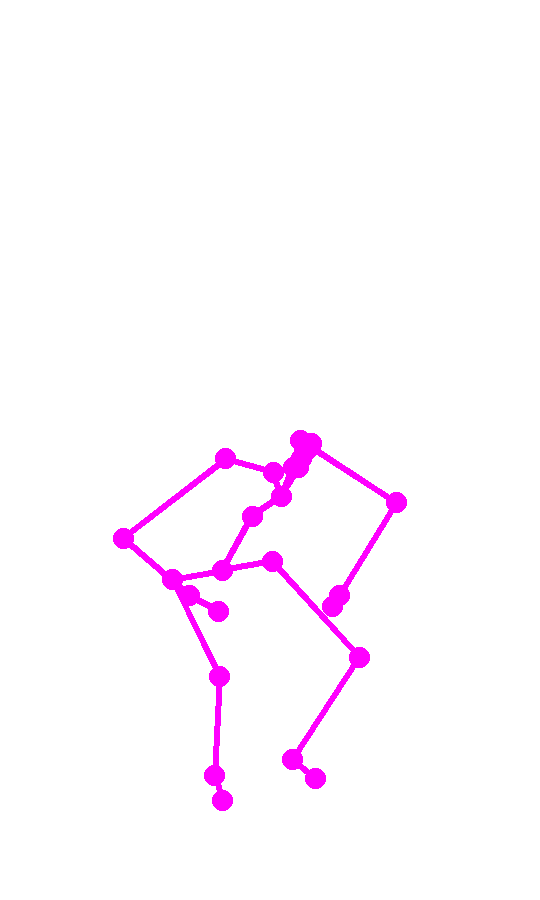}}
	\subfigure[Chair Stands ]{\label{fig:b}
		\includegraphics[trim=0 30 0 100, clip, width=0.15\textwidth]{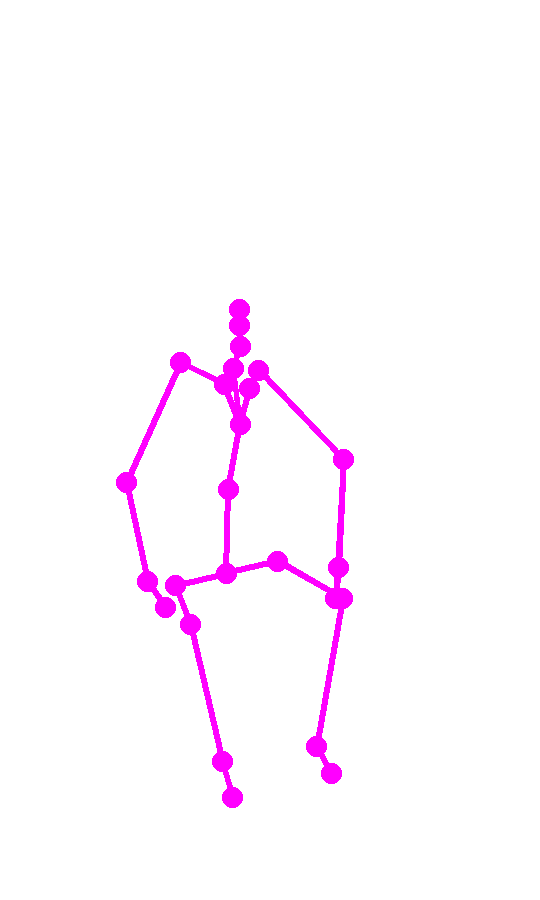}}
	\subfigure[Buddha's Prayer]{\label{fig:c}
		\includegraphics[trim=0 30 0 100, clip, width=0.15\textwidth]{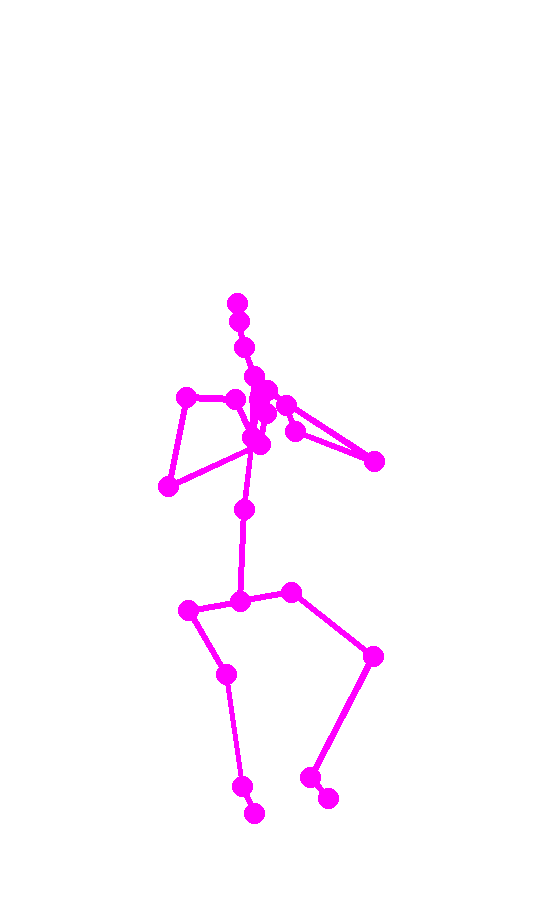}}
	\subfigure[Cops \& Robbers]{\label{fig:d}
		\includegraphics[trim=0 30 0 100, clip, width=0.15\textwidth]{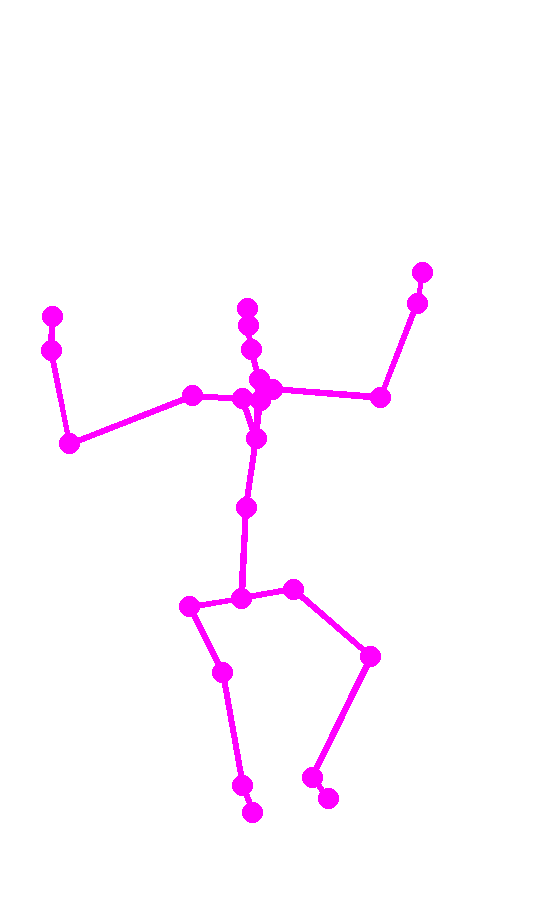}}
	\subfigure[Abs in, Knee Lifts]{\label{fig:e}
		\includegraphics[trim=0 30 0 100, clip, width=0.15\textwidth]{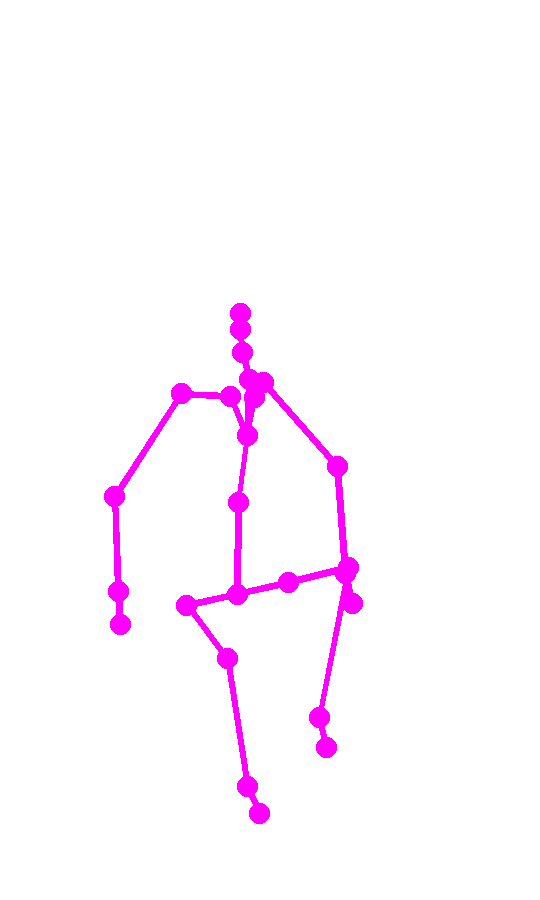}}
	\subfigure[Lateral Stepping]{\label{fig:f}
		\includegraphics[trim=0 30 0 100, clip, width=0.15\textwidth]{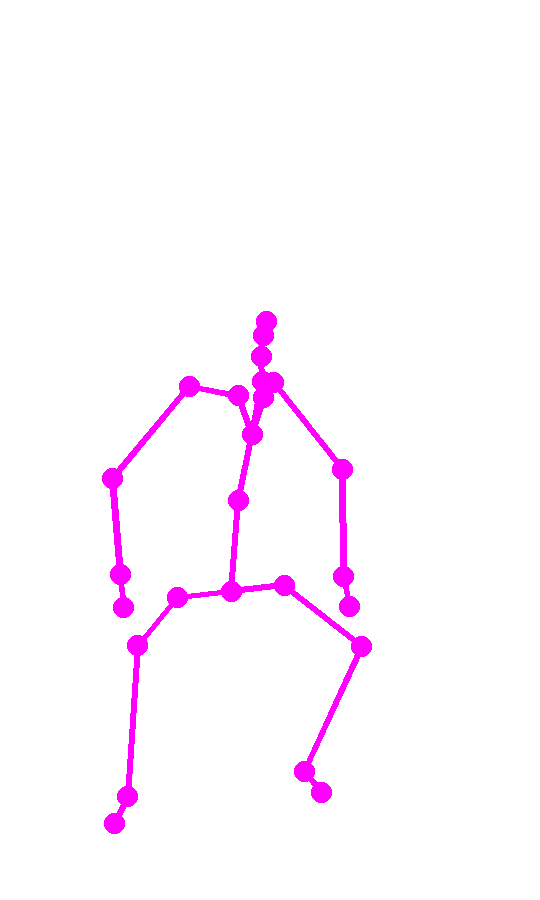}}
	\\
	\subfigure[Pickup \& Throw]{\label{fig:g}
		\includegraphics[trim=0 70 0 60, clip, width=0.15\textwidth]{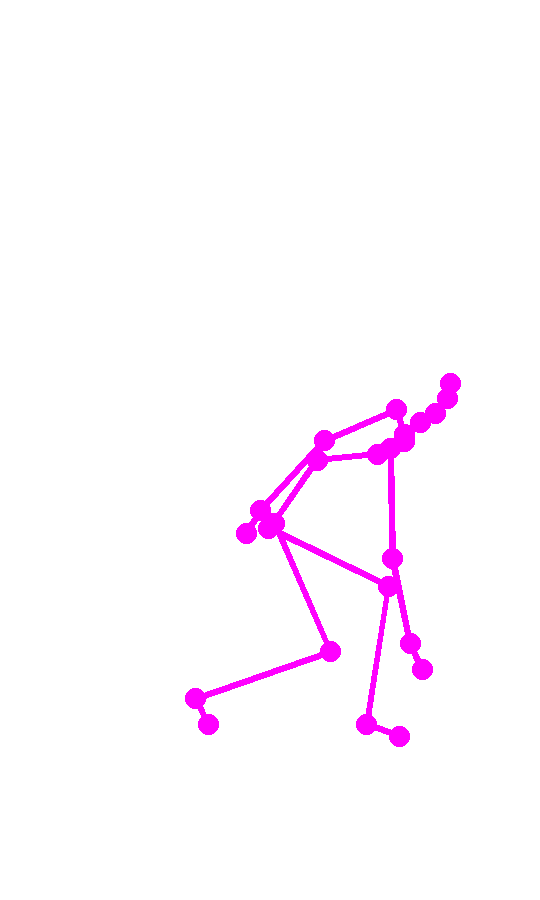}}
	\subfigure[Jogging]{\label{fig:h}
		\includegraphics[trim=0 70 0 60, clip, width=0.15\textwidth]{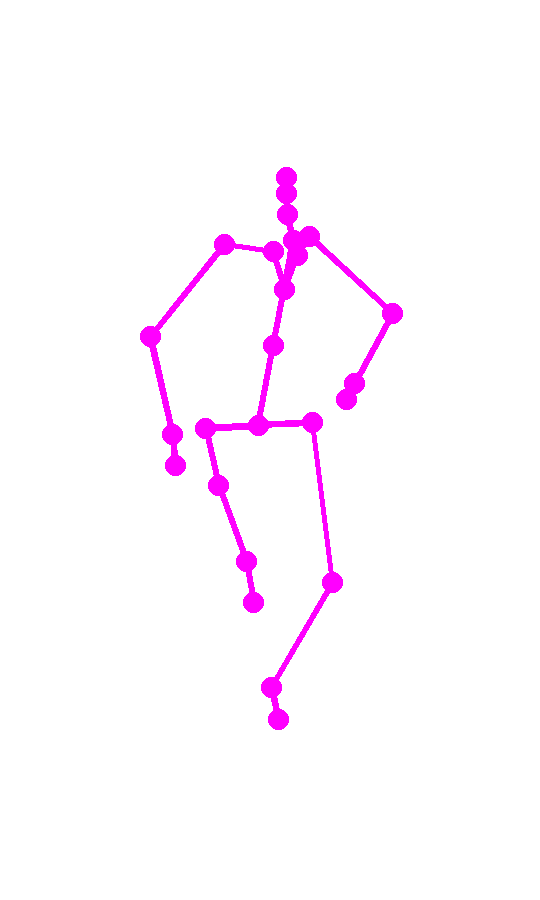}}
	\subfigure[Clapping]{\label{fig:i}
		\includegraphics[trim=0 70 0 60, clip, width=0.15\textwidth]{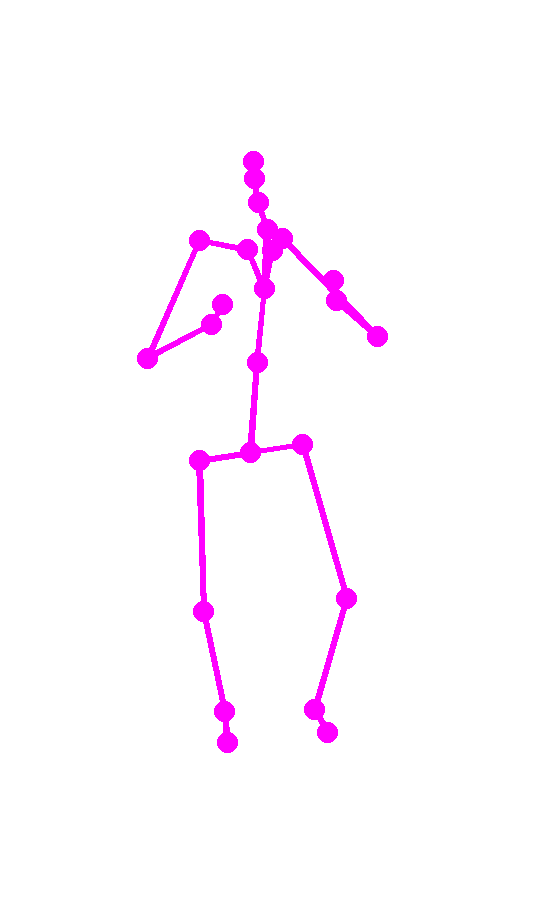}}
	\subfigure[Punching]{\label{fig:j}
		\includegraphics[trim=0 70 0 60, clip, width=0.15\textwidth]{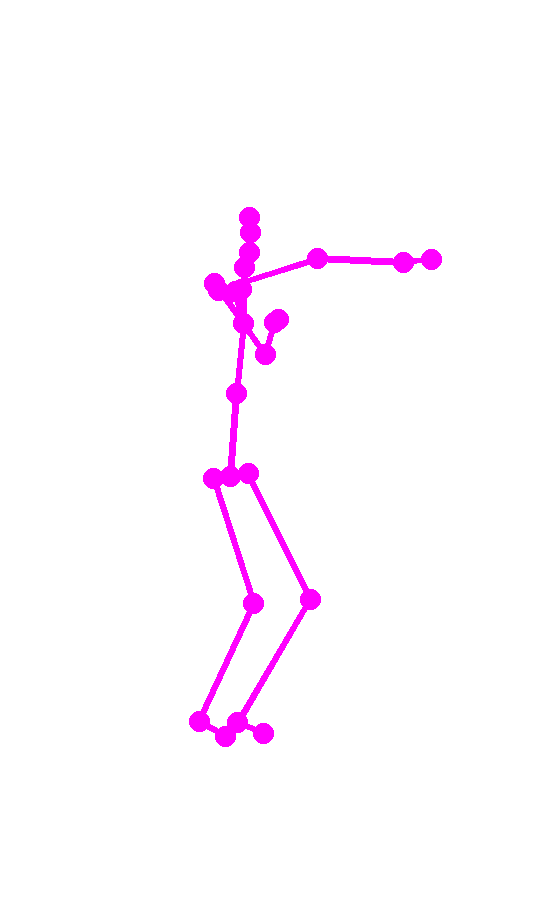}}
	\subfigure[Line Stepping]{\label{fig:k}
		\includegraphics[trim=0 55 0 75, clip, width=0.15\textwidth]{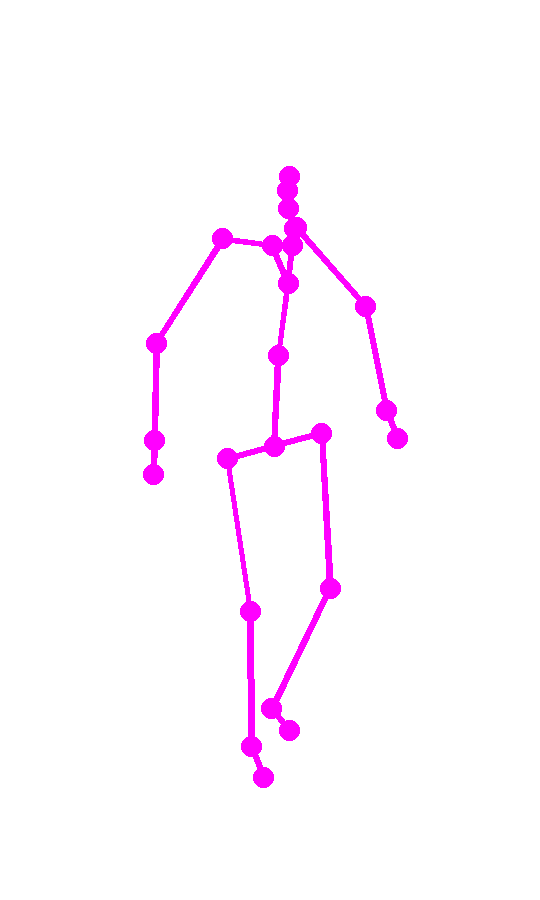}}
	\subfigure[Pendulum]{\label{fig:l}
		\includegraphics[trim=0 60 0 70, clip, width=0.15\textwidth]{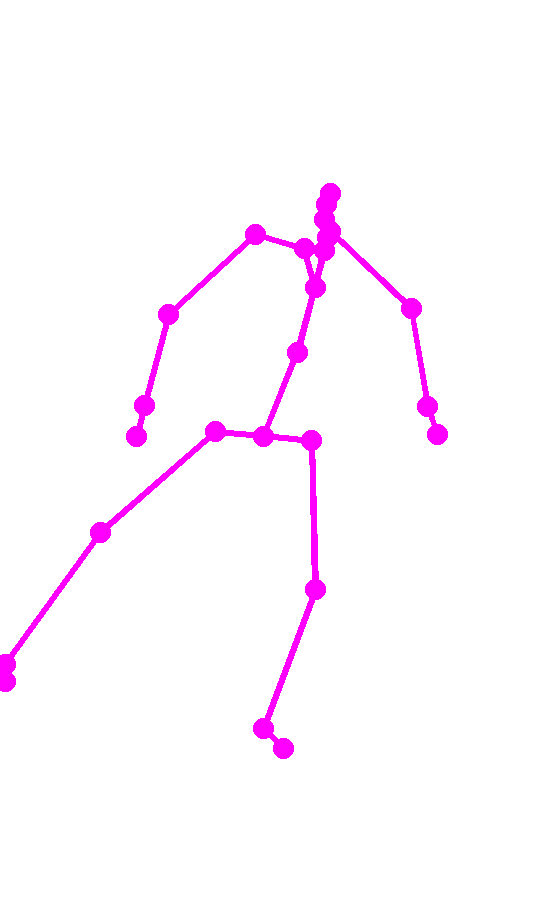}}
	\caption{Motion capture skeleton of the key poses for the 12 exercises shown in Fig.~\ref{fig:exer_snapshots}.}
	\label{fig:exer_snapshots_skeleton}
\end{figure*}

We captured the motion data in 10 subjects (mean age: 27). Before starting the recording, each subject was instructed on how to perform the exercise via a video. We first recorded the motion capture calibration sequence for the subsequent skeleton fitting. Each exercise recording consisted of five repetitions, except for the \emph{Jogging} which required subjects to perform ten jogging steps. The recording of the 12 exercises was repeated for three different orientation angles of the subjects with respect to the Kinect cameras, i.e. at ${0}^\circ$ with subject facing the cameras and at ${30}^\circ$ and ${60}^\circ$ with subject rotated to the left of the cameras. Figs.~\ref{fig:exer_snapshots} and \ref{fig:exer_snapshots_skeleton} show the video snapshots and the corresponding motion capture skeletons of the key poses for the 12 exercises in one of the subjects.

After the data acquisition, the joint coordinates of Kinect~1 and Kinect~2 were transformed into the global coordinate system of the motion capture. Additionally, the temporal data were synchronized according to the time stamp of the sequence captured by Kinect~2, as described in Section~\ref{ssec:Calibration}. 

\begin{figure}[!htbp]
	\begin{center} 
		\includegraphics[width=7cm]{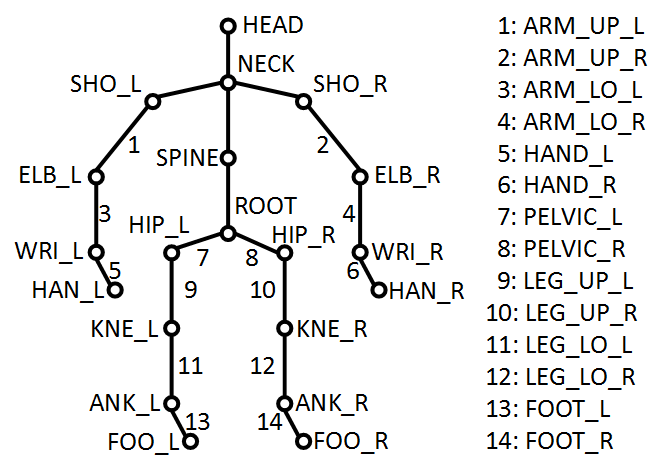}
		\caption{Diagram of the 20-joint skeleton with labeled joint and bone segments that are used in the analysis (L-left, R-right, UP-upper, LO-lower).}
		\label{fig:skeleton}
	\end{center}
\end{figure}

For the analysis of joint position accuracy, we selected 20 joints that are common between the three systems. These joints and their abbreviated names are shown in Fig.~\ref{fig:skeleton}. In addition to the joint position accuracy, we also evaluated the accuracy of the bone lengths for the upper and lower extremities. Those bones and their abbreviated names are also shown in Fig.~\ref{fig:skeleton}.

	\section{Results and Discussion}
	\label{sec:results}
In this section, we present detailed analysis of the pose tracking accuracy in Kinect~1 and Kinect~2 in comparison to the motion capture system which we use as a baseline. All the reported results are the average values across all the subjects. The values in the sitting or standing pose represents the mean values of all the exercises in the sitting or standing set.

\subsection{Joint Position Accuracy}

Tables~\ref{tab:joint_pos_sitting} and \ref{tab:joint_pos_standing} summarize the mean offsets for all the joints in the sitting and standing sets of exercises in three different viewpoint directions. The mean offset represents the average distance between the corresponding joint position of Kinect~1 or Kinect~2 as compared to the location identified from the motion capture data. 

In the sitting set of exercises (Table~\ref{tab:joint_pos_sitting}), the majority of the mean joint offsets range between 50 mm and 100 mm for both Kinect systems. The largest offset in Kinect~1 is consistently observed in the pelvic area which includes the following three joints: ROOT, HIP\_L, and HIP\_R. Kinect~2 on the other hand has smaller offsets for these particular joints. In Kinect~2, the largest offsets are observed in the following four joints of the lower extremities: ANK\_L, ANK\_R, FOO\_L, and FOO\_R. These joints typically have a large vertical offset from the ground plane, while the same is not observed in Kinect~1. Similar observations can be made in the standing set of exercises (Table~\ref{tab:joint_pos_standing}) where the largest offsets in Kinect~1 are again in the pelvic area and the largest offsets in Kinect~2 are found in the lower extremities. These observations are also clearly visible in Fig.~\ref{fig:skeleton_demo}.

Tables~\ref{tab:joint_pos_sitting} and \ref{tab:joint_pos_standing} also summarize the standard deviation (SD) of the joint position offsets which reflects the variability of a particular joint tracking. For most of the joints, the SD ranges between 10 mm and 50 mm. The joints that exhibit considerable motion during an exercise have much higher variability, and thus SD, typically greater than 50 mm. In most cases, the SDs of the joint positions in Kinect~2 are considerably smaller than those in Kinect~1. This is most likely due to an increased resolution and reduced noise level of Kinect~2 depth maps.

Furthermore, we can observe that the mean offset and SD of the joints that are most active in a particular exercise are both increasing with the viewpoint angle. This is especially noticeable on the side of the skeleton that is turned further away from the camera as the occlusion of joints increases the uncertainty of the pose detection. In our experiments, the left side of the skeleton was turning away from the camera with the increasing viewpoint angle.

\begin{figure*}[htbp] 
	\begin{minipage}[t]{0.48\linewidth}
		\begin{tabular}{ccc}
			\multicolumn{3}{c}{\includegraphics[width = 4cm]{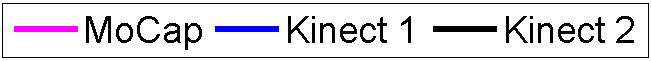}}\\
			\includegraphics[trim=0 40 0 100, clip, width=2.7cm] {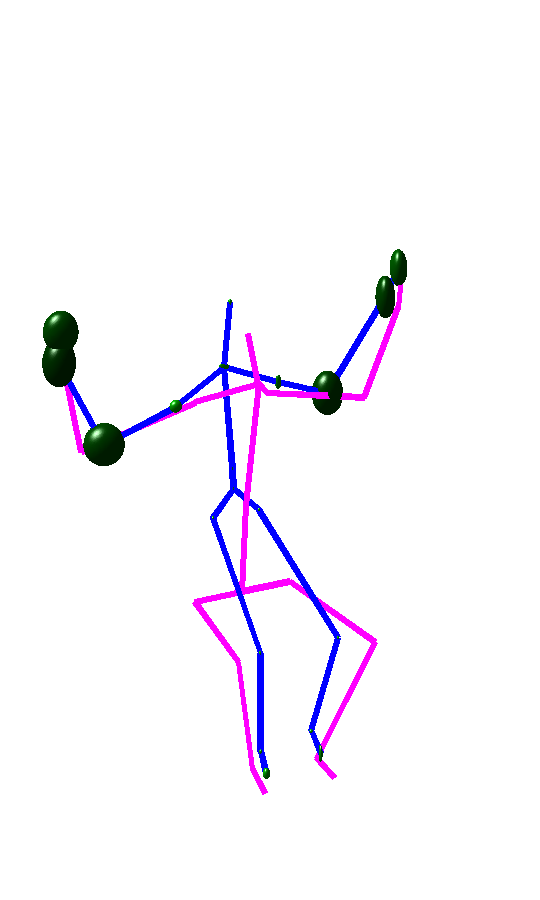}&
			\includegraphics[trim=0 40 0 100, clip, width=2.7cm] {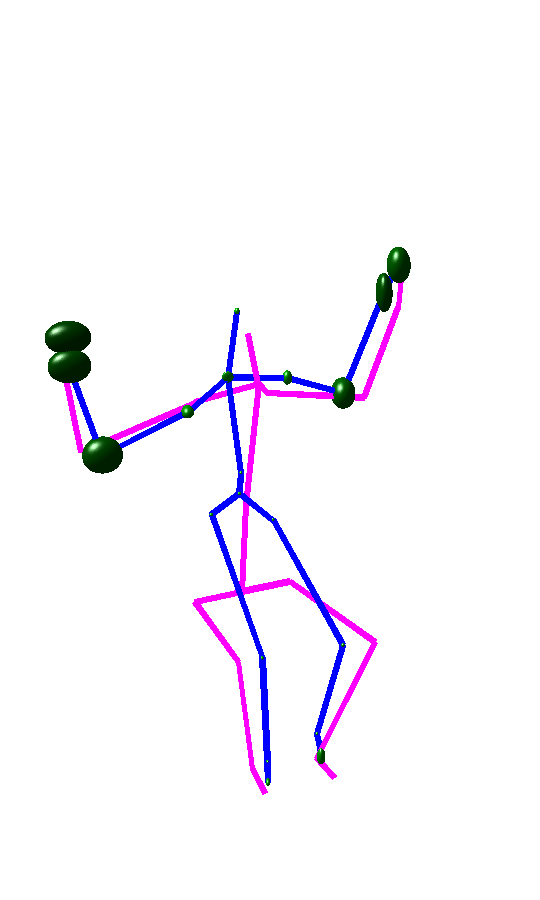}&
			\includegraphics[trim=0 40 0 100, clip, width=2.7cm] {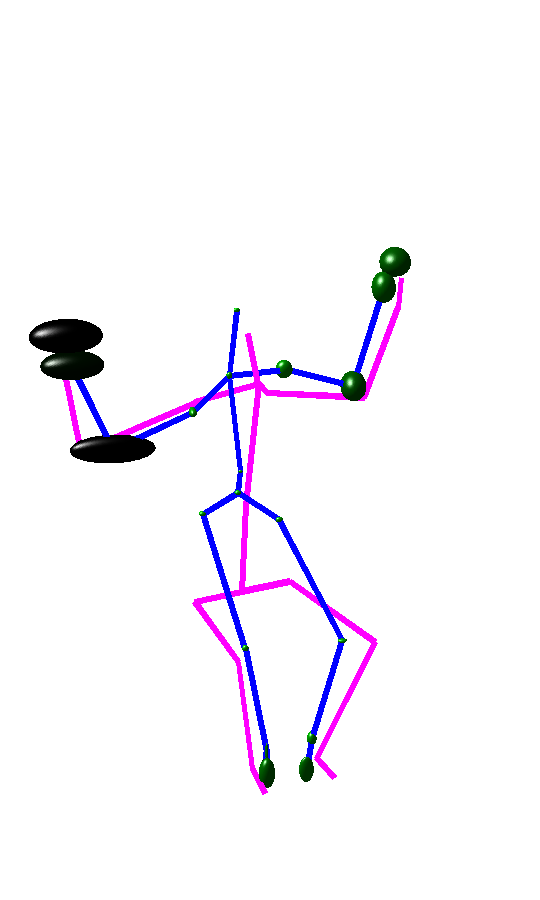}\\
			0$^{\circ}$  & 30$^{\circ}$  & 60$^{\circ}$   \\
			\includegraphics[trim=0 40 0 110, clip, width=2.7cm] {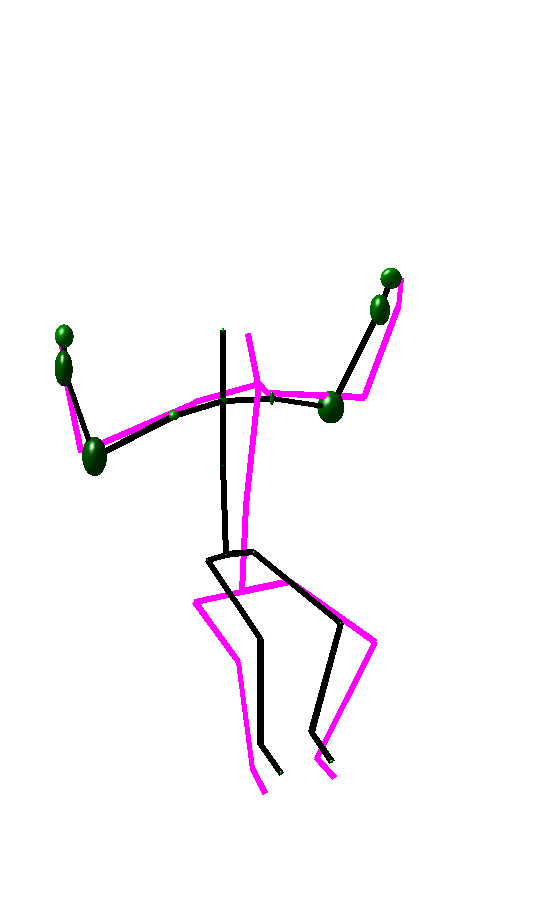}&
			\includegraphics[trim=0 40 0 110, clip, width=2.7cm] {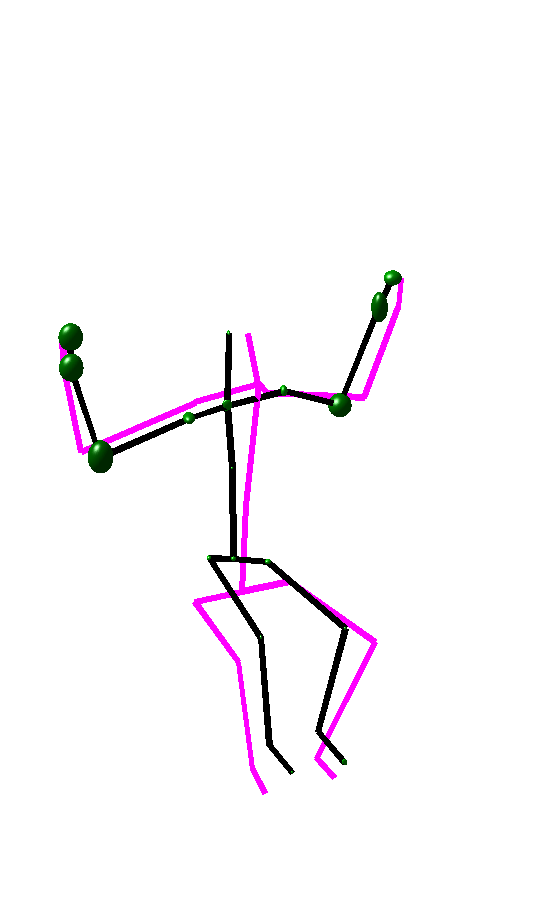}&
			\includegraphics[trim=0 40 0 110, clip, width=2.7cm] {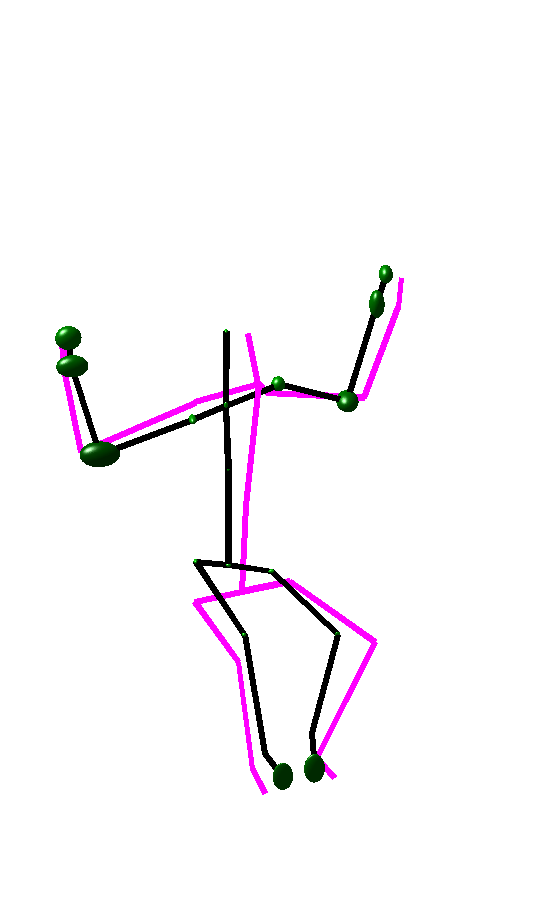}\\
			0$^{\circ}$ & 30$^{\circ}$ & 60$^{\circ}$ \\
		\end{tabular}
		\caption{Mean offset and SD of skeletal joints for the exercise \emph{Cops \& Robbers} (top row: Kinect~1, bottom row: Kinect~2) as captured at three different viewpoint angles.}
		\label{fig:joint_mean_sd_1}
	\end{minipage}
	\hfill
	\begin{minipage}[t]{0.48\linewidth}
		\begin{tabular}{ccc}
			\multicolumn{3}{c}{\includegraphics[width = 4cm]{legend.png}}\\
			\includegraphics[trim=0 40 0 100, clip, width=2.7cm] {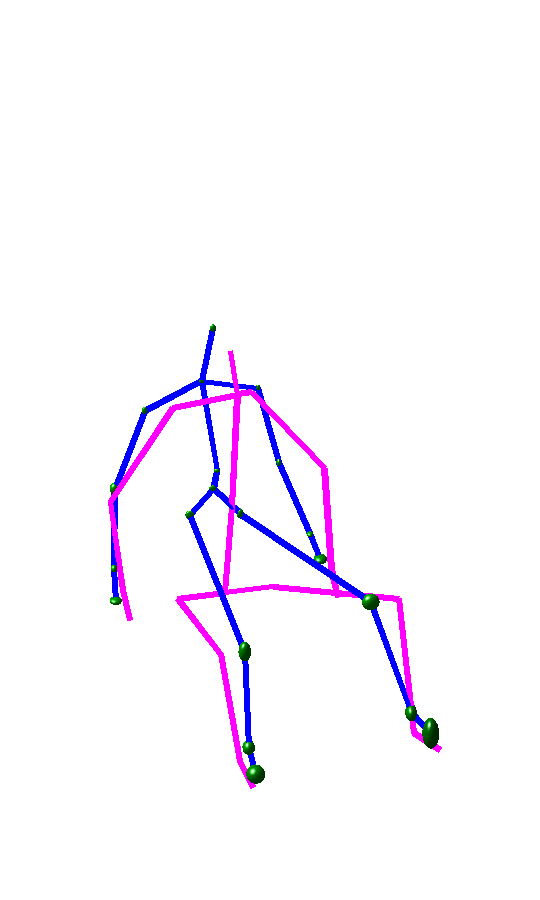}&
			\includegraphics[trim=0 40 0 100, clip, width=2.7cm] {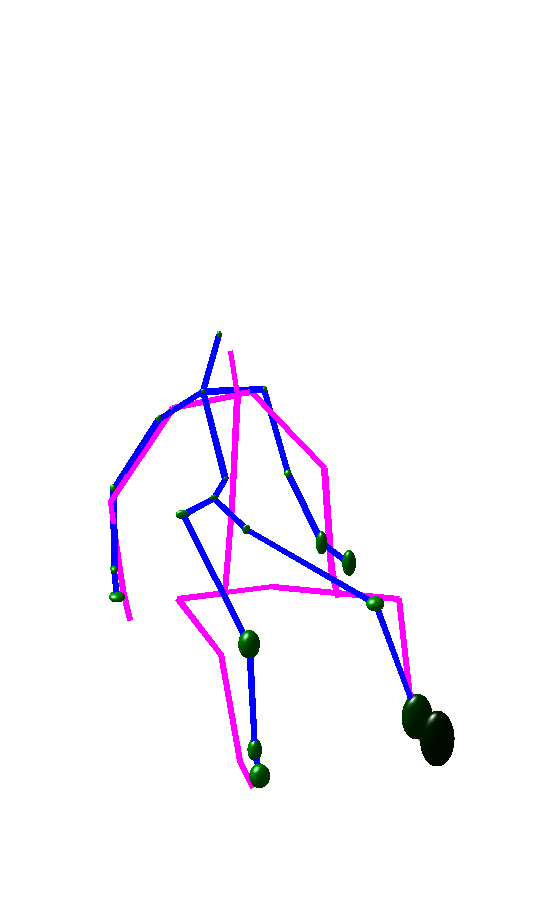}&
			\includegraphics[trim=0 40 0 100, clip, width=2.7cm] {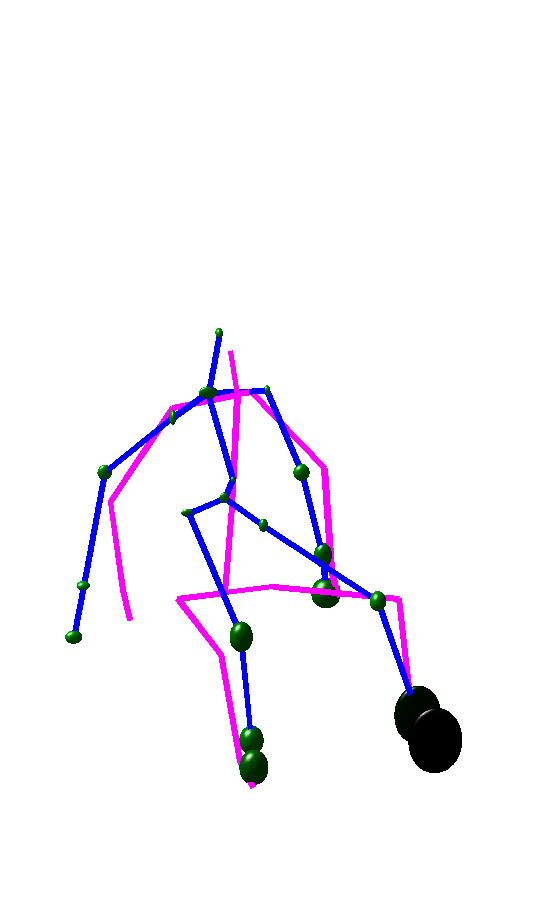}\\
			0$^{\circ}$ & 30$^{\circ}$ & 60$^{\circ}$ \\
			\includegraphics[trim=0 40 0 100, clip, width=2.7cm] {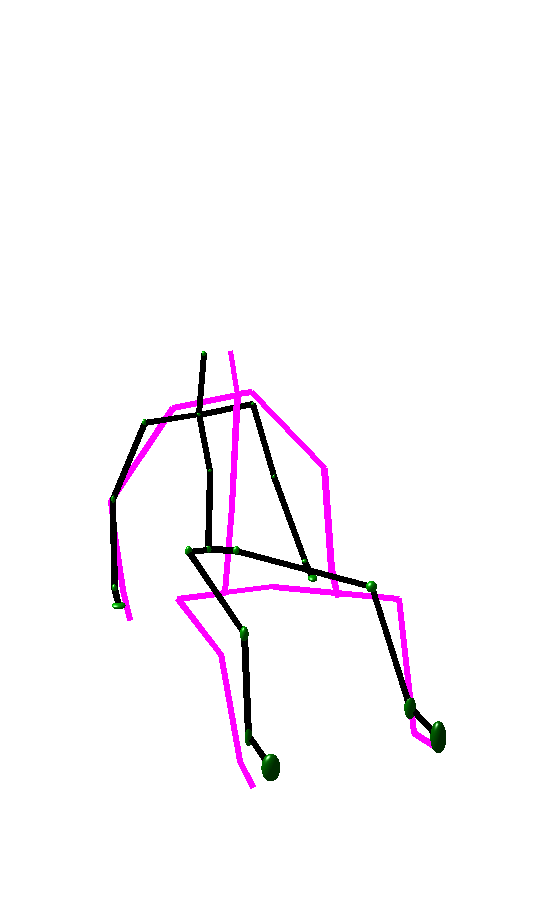}&
			\includegraphics[trim=0 40 0 100, clip, width=2.7cm] {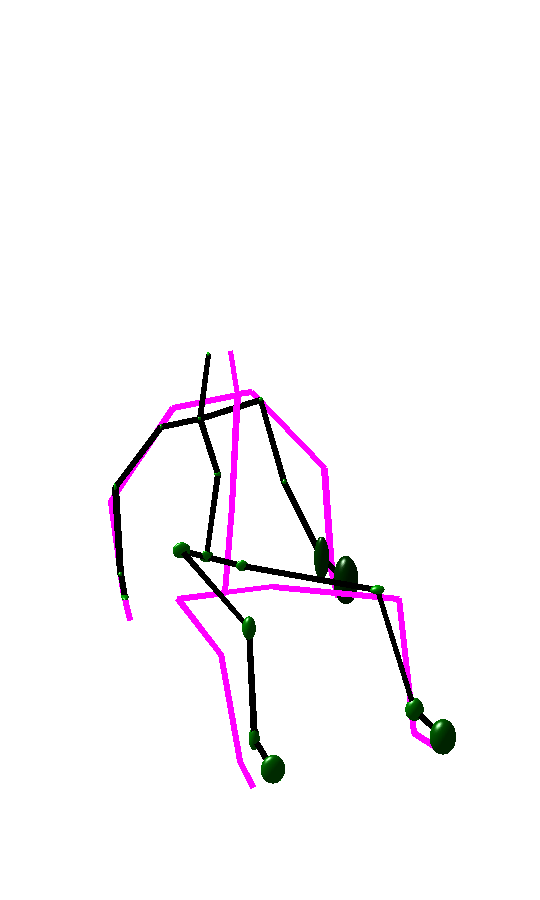}&
			\includegraphics[trim=0 40 0 100, clip, width=2.7cm] {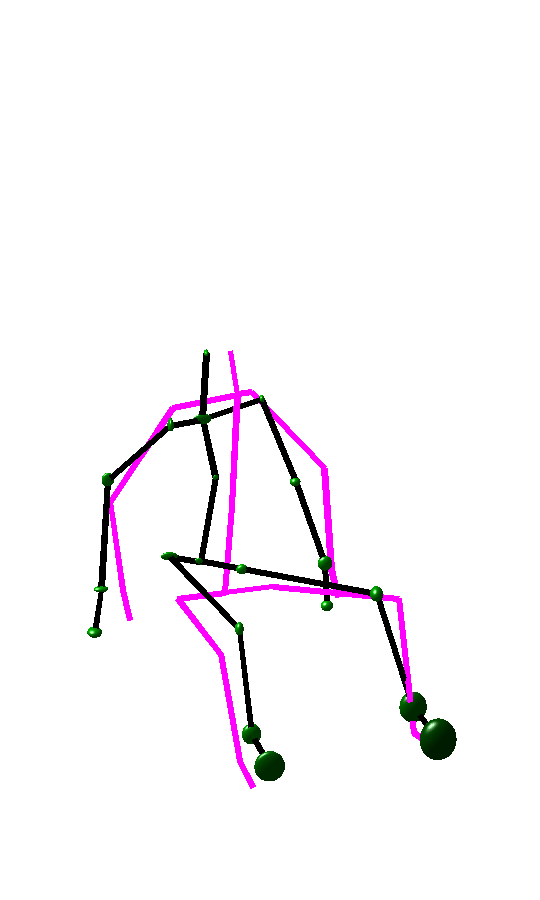}\\
			0$^{\circ}$ & 30$^{\circ}$ & 60$^{\circ}$ \\
		\end{tabular}
		\caption{Mean offset and SD of skeletal joints for the exercise \emph{Lateral Stepping} (top row: Kinect~1, bottom row: Kinect~2) as captured at three different viewpoint angles.}
		\label{fig:joint_mean_sd_2}
	\end{minipage}
\end{figure*}
\begin{figure*}[htbp] 
	\begin{minipage}[t]{0.48\linewidth}
		\begin{tabular}{ccc}
			\multicolumn{3}{c}{\includegraphics[width = 4cm]{legend.png}}\\
			\includegraphics[trim=0 80 0 60, clip, width=2.7cm] {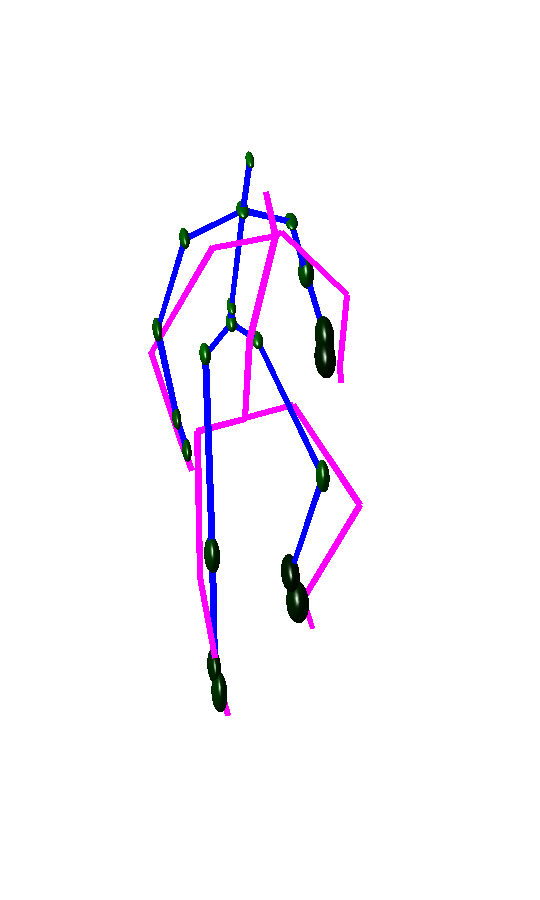}&
			\includegraphics[trim=0 80 0 60, clip, width=2.7cm] {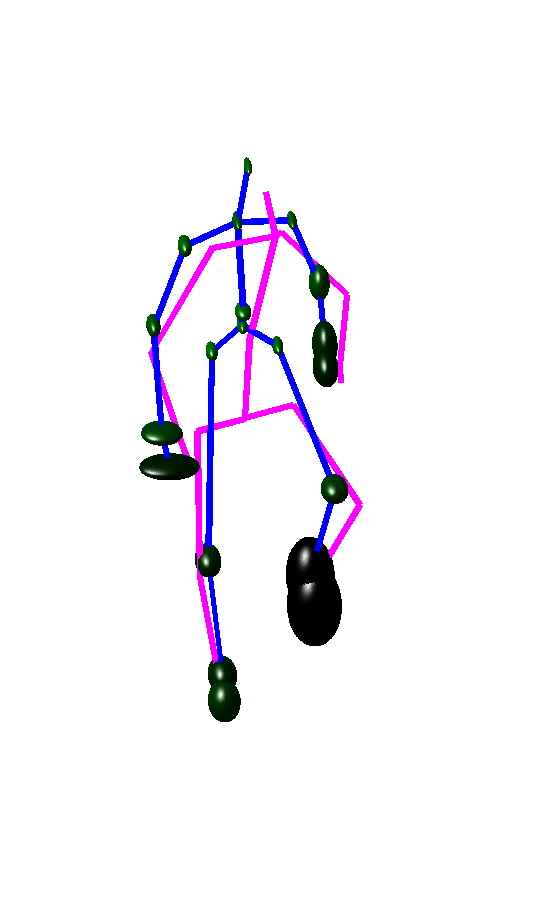}&
			\includegraphics[trim=0 80 0 60, clip, width=2.7cm] {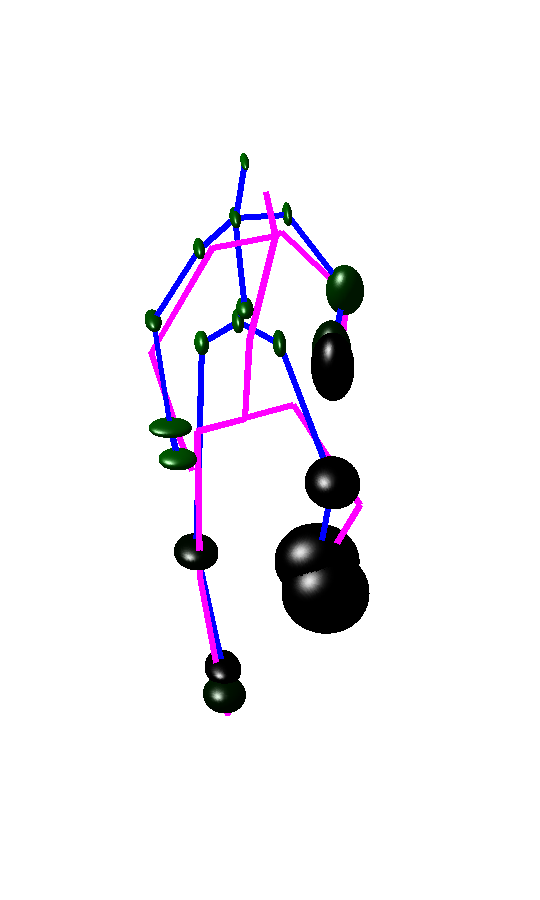}\\
			0$^{\circ}$ & 30$^{\circ}$ &60$^{\circ}$ \\
			\includegraphics[trim=0 80 0 70, clip, width=2.7cm] {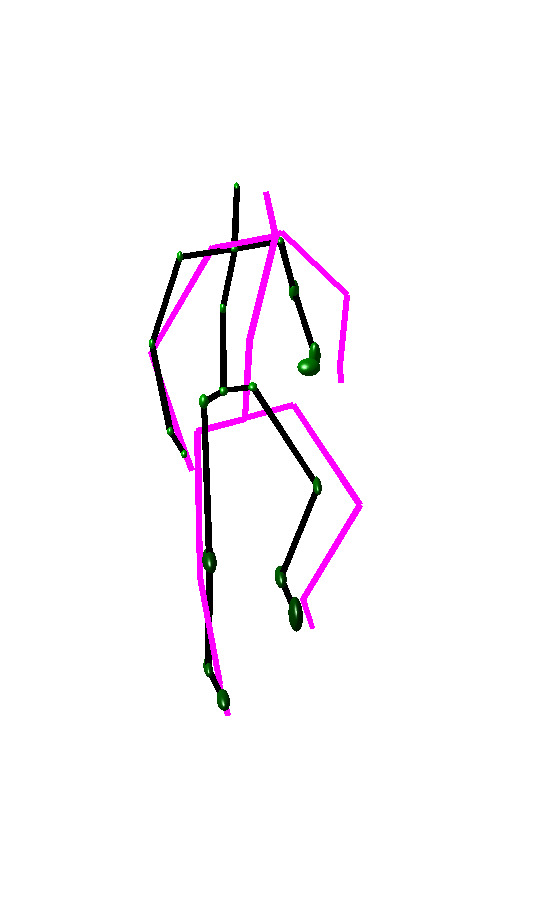}&
			\includegraphics[trim=0 80 0 70, clip, width=2.7cm] {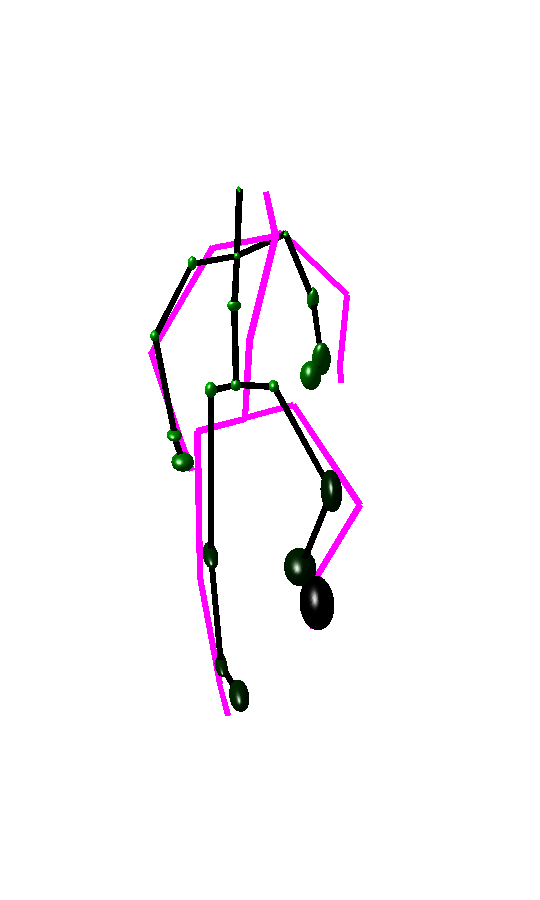}&
			\includegraphics[trim=0 80 0 70, clip, width=2.7cm] {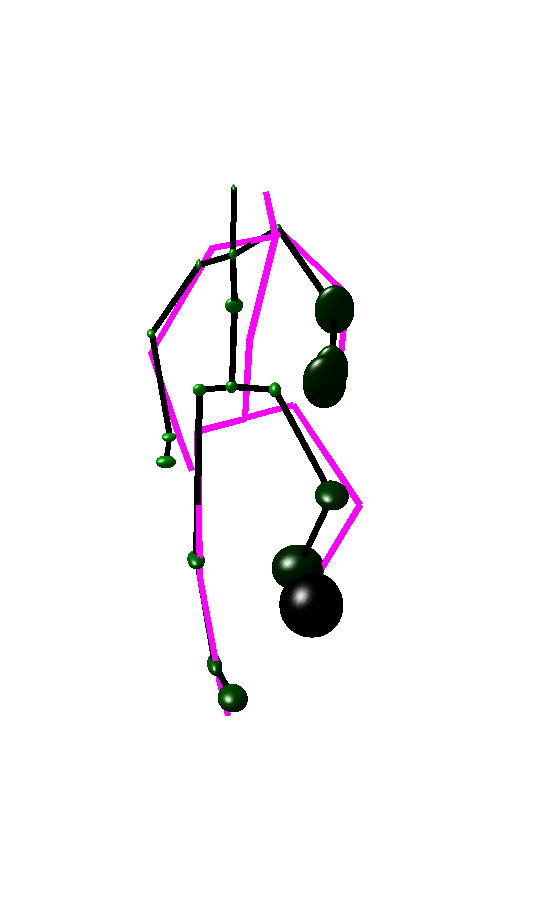}\\
			0$^{\circ}$ & 30$^{\circ}$ &60$^{\circ}$\\
		\end{tabular}
		\caption{Mean offset and SD of skeletal joints for the exercise \emph{Jogging} (top row: Kinect~1, bottom row: Kinect~2) as captured at three different viewpoint angles.}
		\label{fig:joint_mean_sd_3}
	\end{minipage}
	\hfill
	\begin{minipage}[t]{0.48\linewidth}
		\begin{tabular}{ccc}
			\multicolumn{3}{c}{\includegraphics[width = 4cm]{legend.png}}\\
			\includegraphics[trim=0 70 0 70, clip, width=2.7cm] {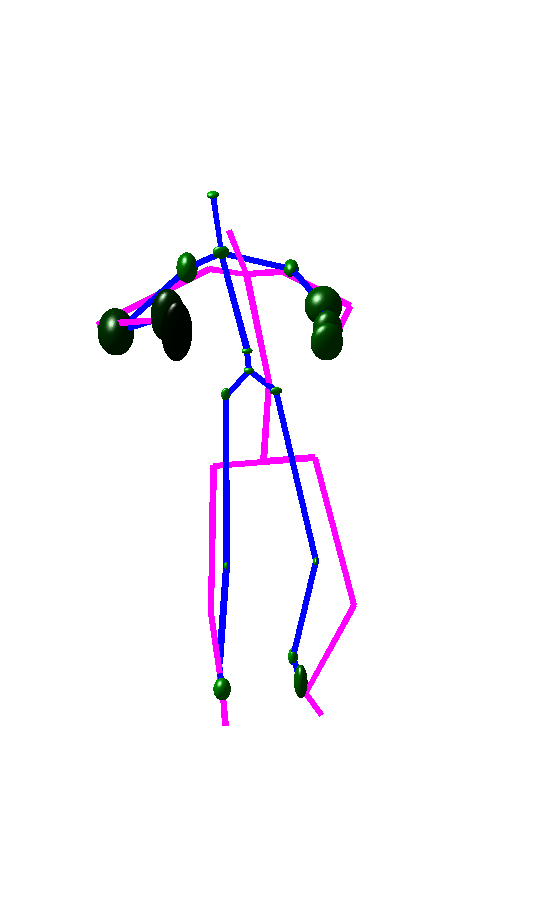}&
			\includegraphics[trim=0 70 0 70, clip, width=2.7cm] {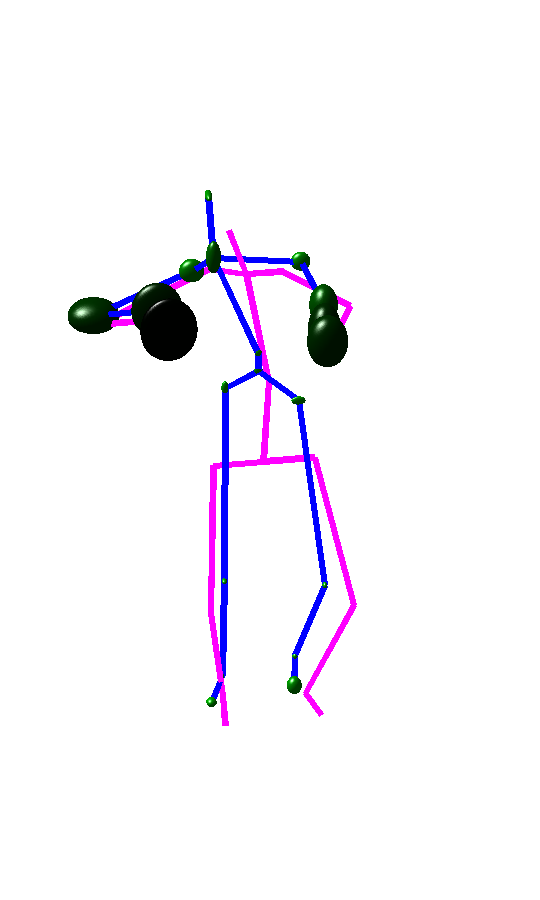}&
			\includegraphics[trim=0 70 0 70, clip, width=2.7cm] {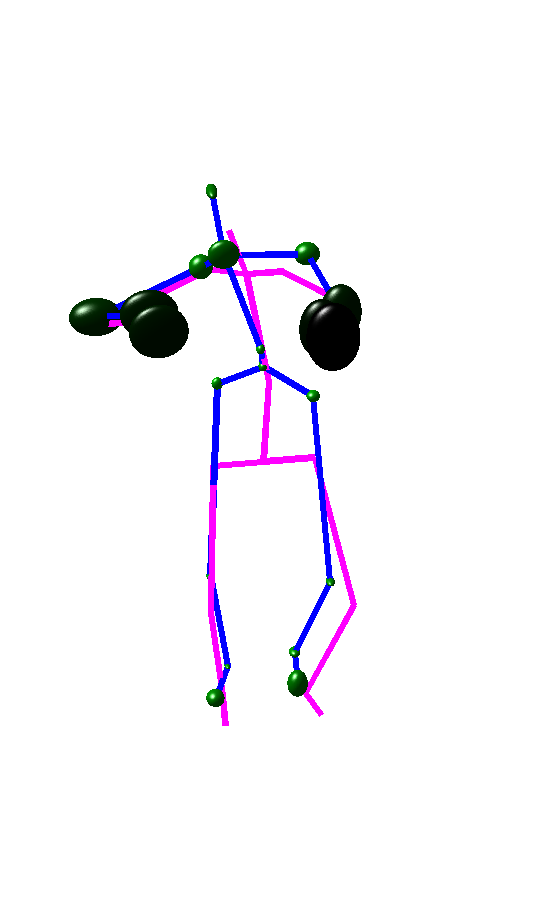}\\
			0$^{\circ}$ & 30$^{\circ}$ & 60$^{\circ}$ \\
			\includegraphics[trim=0 70 0 80, clip, width=2.7cm] {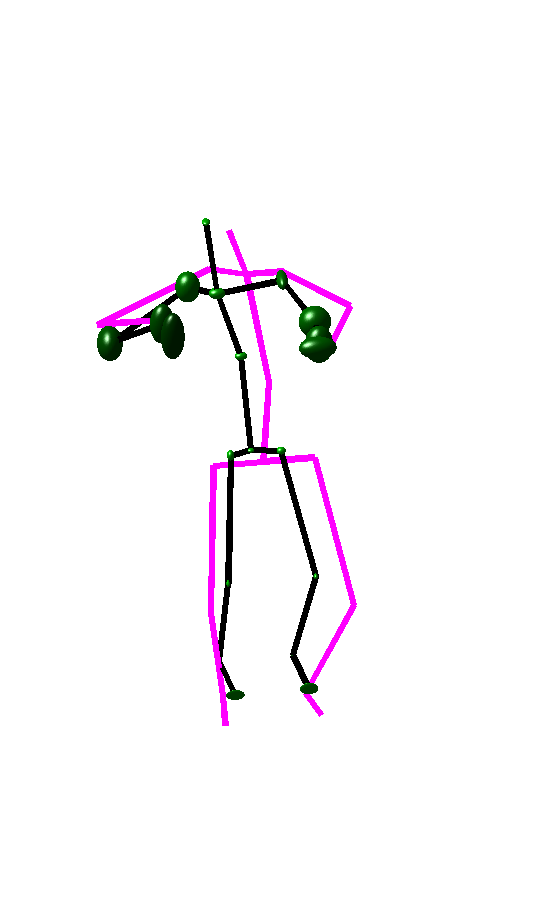}&
			\includegraphics[trim=0 70 0 80, clip, width=2.7cm] {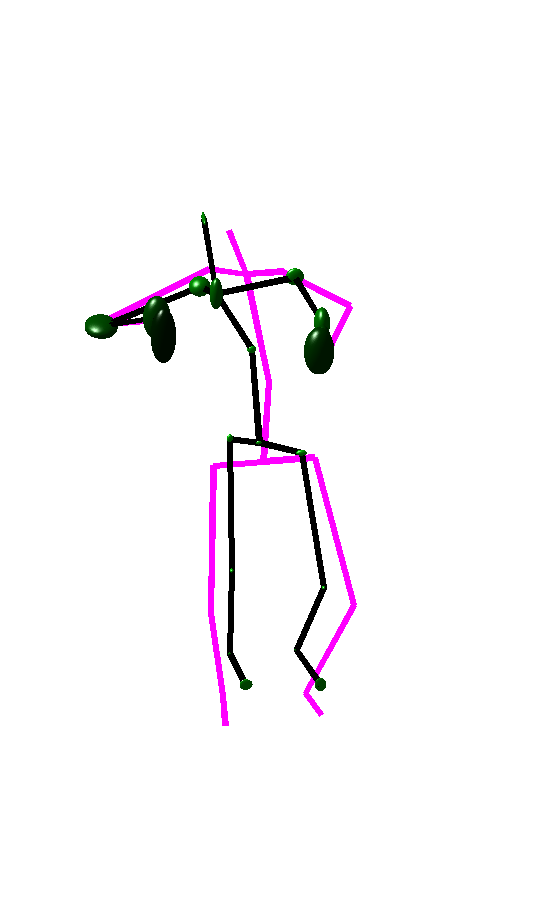}&
			\includegraphics[trim=0 70 0 80, clip, width=2.7cm] {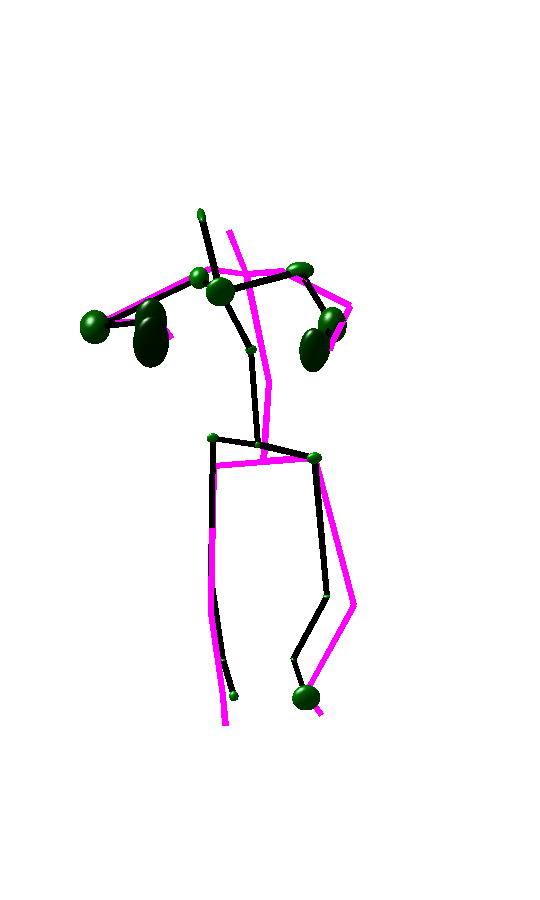}\\
			0$^{\circ}$ & 30$^{\circ}$ &60$^{\circ}$ \\
		\end{tabular}
		\caption{Mean offset and SD of skeletal joints for the exercise \emph{Punching} (top row: Kinect~1, bottom row: Kinect~2) as captured at three different viewpoint angles.}
		\label{fig:joint_mean_sd_4}
	\end{minipage}
\end{figure*}

In order to examine whether Kinect~2 is performing better than Kinect~1, we performed statistical analysis of the accuracy for each of the 20 joints in comparison to the motion capture. We used pair-wise \textit{t}-test for the joint position analysis. Our hypothesis was that the position of a particular joint is not significantly different between a Kinect and the motion capture. The results of the analysis are shown in Tables \ref{tab:joint_pos_sitting} and \ref{tab:joint_pos_standing} where the joints with significant difference are denoted with * symbol when p $<$ 0.05 and ** symbol when p $<$ 0.01, respectively.

The results of the \textit{t}-test analyses show that the joint position offsets of the joints ROOT, SPINE, HIP\_L, HIP\_R, ANK\_L, ANK\_R, FOO\_L, are FOO\_R for Kinect~2 have significantly different mean offsets as compared to Kinect~1. The mean joint position offsets of other joints share the same distribution.

Similar conclusions can be drawn for the standing set of exercises (Table \ref{tab:joint_pos_standing}). For example, the SDs of the joint position offset in Kinect~2 are usually smaller than those of Kinect~1. The variances in the more active joints are typically increasing with the viewpoint angle. Statistically significant differences in the accuracy of Kinect~1 vs. Kinect~2 can be found in the following joints: ROOT, SPINE, HIP\_L, HIP\_R, ANK\_L, ANK\_R, FOO\_L, and FOO\_R. Overall, the means and SDs of the joint position offsets in the standing poses are usually larger than those in the sitting poses. In the sitting poses there are higher number of static joints, which in general have smaller variability.

\begin{table*}[t!]
	\centering
	\hspace{-7mm}
	\begin{minipage}[t]{0.45\linewidth}
		\centering
		\caption{Joint position offsets in sitting exercises.}
		\label{tab:joint_pos_sitting}%
		\begin{tabular}{|l|p{0.2cm}|p{0.2cm}|p{0.2cm}|p{0.2cm}|p{0.2cm}|p{0.2cm}||p{0.2cm}|p{0.2cm}|p{0.2cm}|p{0.2cm}|p{0.2cm}|p{0.2cm}|}
			\hline
			\multirow{3}{*}{} & \multicolumn{6}{c||}{Kinect~1}                  & \multicolumn{6}{c|}{Kinect~2} \\
			\cline{2-13}
			& \multicolumn{3}{c|}{Mean (mm)} & \multicolumn{3}{c||}{SD (mm)} & \multicolumn{3}{c|}{Mean (mm)} & \multicolumn{3}{c|}{SD (mm)} \\
			\cline{2-13}
			& 0$^{\circ}$     & 30$^{\circ}$    & 60$^{\circ}$    & 0$^{\circ}$     & 30$^{\circ}$    & 60$^{\circ}$    & 0$^{\circ}$     & 30$^{\circ}$    & 60$^{\circ}$    & 0$^{\circ}$     & 30$^{\circ}$    & 60$^{\circ}$ \\
			\hline
			ROOT$**$\footnotetext[0]{* \textit{t}-test, p $<$ 0.05, ** \textit{t}-test, p $<$ 0.01}  & 256   & 262   & 263   & 25    & 20    & 25    & 100   & 102   & 106   & 17    & 18    & 16 \\
			SPINE & 91    & 97    & 100   & 24    & 20    & 21    & 110   & 117   & 126   & 13    & 14    & 11 \\
			NECK  & 79    & 65    & 62    & 25    & 21    & 23    & 84    & 78    & 73    & 14    & 16    & 14 \\
			HEAD  & 74    & 70    & 67    & 26    & 21    & 21    & 50    & 51    & 50    & 13    & 15    & 12 \\
			SHO\_L & 90    & 89    & 97    & 26    & 24    & 33    & 76    & 76    & 82    & 16    & 24    & 29 \\
			ELB\_L & 81    & 86    & 98    & 27    & 29    & 35    & 87    & 103   & 114   & 17    & 28    & 25 \\
			WRI\_L & 76    & 90    & 118   & 33    & 48    & 55    & 59    & 84    & 115   & 25    & 53    & 44 \\
			HAN\_L & 85    & 106   & 134   & 41    & 60    & 68    & 64    & 95    & 125   & 31    & 60    & 53 \\
			SHO\_R & 78    & 74    & 69    & 28    & 24    & 23    & 80    & 83    & 78    & 17    & 20    & 17 \\
			ELB\_R & 95    & 93    & 89    & 28    & 38    & 34    & 88    & 77    & 70    & 21    & 25    & 19 \\
			WRI\_R & 64    & 93    & 110   & 30    & 53    & 65    & 61    & 64    & 71    & 25    & 28    & 24 \\
			HAN\_R & 83    & 113   & 130   & 35    & 65    & 80    & 74    & 71    & 75    & 24    & 28    & 26 \\
			HIP\_L$**$ & 188   & 200   & 215   & 25    & 20    & 22    & 115   & 122   & 139   & 16    & 17    & 15 \\
			KNE\_L & 96    & 95    & 93    & 24    & 24    & 27    & 76    & 95    & 114   & 16    & 18    & 25 \\
			ANK\_L$*$ & 54    & 77    & 81    & 17    & 25    & 32    & 93    & 103   & 113   & 16    & 18    & 29 \\
			FOO\_L$*$ & 66    & 74    & 86    & 19    & 26    & 38    & 93    & 105   & 119   & 18    & 25    & 39 \\
			HIP\_R$**$ & 207   & 210   & 211   & 25    & 21    & 24    & 133   & 128   & 132   & 15    & 18    & 16 \\
			KNE\_R & 123   & 118   & 128   & 23    & 23    & 34    & 120   & 117   & 119   & 15    & 17    & 24 \\
			ANK\_R$*$ & 67    & 75    & 91    & 17    & 21    & 27    & 112   & 122   & 132   & 15    & 19    & 27 \\
			FOO\_R$*$ & 67    & 74    & 84    & 17    & 20    & 26    & 126   & 134   & 139   & 19    & 20    & 28 \\
			\hline
		\end{tabular}%
	\end{minipage}
	\hspace{10mm}
	\begin{minipage}[t]{0.45\linewidth}
		\centering
		\caption{Joint position offsets in standing exercises.}
		\label{tab:joint_pos_standing}%
		\begin{tabular}{|l|p{0.2cm}|p{0.2cm}|p{0.2cm}|p{0.2cm}|p{0.2cm}|p{0.2cm}||p{0.2cm}|p{0.2cm}|p{0.2cm}|p{0.2cm}|p{0.2cm}|p{0.2cm}|}
			\hline
			\multirow{3}{*}{} & \multicolumn{6}{c||}{Kinect~1}                  & \multicolumn{6}{c|}{Kinect~2} \\
			\cline{2-13}
			& \multicolumn{3}{c|}{Mean (mm)} & \multicolumn{3}{c||}{SD (mm)} & \multicolumn{3}{c|}{Mean (mm)} & \multicolumn{3}{c|}{SD (mm)} \\
			\cline{2-13}
			& 0$^{\circ}$     & 30$^{\circ}$    & 60$^{\circ}$    & 0$^{\circ}$     & 30$^{\circ}$    & 60$^{\circ}$    & 0$^{\circ}$     & 30$^{\circ}$    & 60$^{\circ}$    & 0$^{\circ}$     & 30$^{\circ}$    & 60$^{\circ}$ \\
			\hline
			ROOT$**$\footnotetext[0]{* \textit{t}-test, p $<$ 0.05, ** \textit{t}-test, p $<$ 0.01}  
			& 245   & 256   & 267   & 23    & 25    & 25    & 76    & 81    & 93    & 24    & 19    & 18\\
			SPINE$*$ & 79    & 89    & 102   & 22    & 22    & 24    & 112   & 129   & 144   & 17    & 16    & 17 \\
			NECK  & 82    & 91    & 102   & 23    & 24    & 23    & 113   & 129   & 143   & 18    & 16    & 16 \\
			HEAD  & 89    & 84    & 87    & 30    & 29    & 26    & 79    & 82    & 90    & 26    & 22    & 20 \\
			SHO\_L & 76    & 76    & 80    & 33    & 36    & 38    & 68    & 72    & 78    & 29    & 31    & 31 \\
			ELB\_L & 98    & 112   & 134   & 46    & 52    & 66    & 112   & 137   & 159   & 37    & 41    & 57 \\
			WRI\_L & 85    & 93    & 110   & 56    & 57    & 71    & 66    & 84    & 111   & 47    & 52    & 73 \\
			HAN\_L & 85    & 96    & 114   & 62    & 65    & 80    & 77    & 94    & 121   & 49    & 55    & 80 \\
			SHO\_R & 76    & 76    & 74    & 30    & 30    & 26    & 96    & 106   & 109   & 26    & 24    & 22 \\
			ELB\_R & 98    & 89    & 80    & 38    & 39    & 34    & 96    & 91    & 80    & 33    & 32    & 31 \\
			WRI\_R & 80    & 82    & 80    & 43    & 43    & 38    & 78    & 88    & 87    & 33    & 32    & 31 \\
			HAN\_R & 85    & 83    & 77    & 46    & 49    & 44    & 81    & 79    & 74    & 33    & 34    & 32 \\
			HIP\_L$**$ & 186   & 205   & 228   & 28    & 27    & 27    & 106   & 125   & 150   & 23    & 19    & 21 \\
			KNE\_L & 101   & 111   & 124   & 33    & 38    & 52    & 107   & 129   & 148   & 26    & 34    & 51 \\
			ANK\_L$*$ & 119   & 144   & 135   & 43    & 61    & 58    & 135   & 168   & 174   & 33    & 51    & 58 \\
			FOO\_L$*$ & 115   & 135   & 122   & 42    & 60    & 59    & 130   & 170   & 180   & 33    & 56    & 63 \\
			HIP\_R$**$ & 188   & 195   & 207   & 25    & 24    & 23    & 103   & 104   & 109   & 25    & 17    & 17 \\
			KNE\_R & 105   & 102   & 101   & 31    & 29    & 26    & 115   & 119   & 118   & 26    & 25    & 27 \\
			ANK\_R$*$ & 111   & 113   & 108   & 38    & 39    & 28    & 146   & 158   & 157   & 32    & 33    & 35 \\
			FOO\_R$*$ & 97    & 93    & 86    & 36    & 37    & 29    & 146   & 157   & 150   & 32    & 33    & 38 \\
			\hline
		\end{tabular}%
	\end{minipage}
\end{table*}

Figs. \ref{fig:joint_mean_sd_1}, \ref{fig:joint_mean_sd_2}, \ref{fig:joint_mean_sd_3}, and \ref{fig:joint_mean_sd_4} demonstrate the means and SDs of the joint position offsets  for the exercises \emph{Cops \& Robbers}, \emph{Lateral Stepping}, \emph{Jogging}, and \emph{Punching}, respectively. In the figures, the skeleton in magenta represents one of the key poses in the exercise sequence as captured by the motion capture system. The blue or black lines on the other hand represent the corresponding skeletons generated from the mean joint position offsets that were observed in either Kinect~1 or Kinect~2, respectively. The ellipsoids at each joint denote the SDs in the 3D space, analyzed for each axis of the local coordinate system attached to the corresponding segment.  The larger size of the ellipsoid indicates larger SD of the joint position in Kinect compared with the joint position captured by the motion capture.

Such visualization of results provides a quick and intuitive way for comparison of accuracy of different joints. The results show that the overall SDs are larger in Kinect~1 as compared to Kinect~2. The variability of offsets is also increasing with the viewpoint angle. In more dynamic movements, such as \emph{Jogging} (Fig.~\ref{fig:joint_mean_sd_3}) and \emph{Punching} (Fig.~\ref{fig:joint_mean_sd_4}), the end-points, such as feet and hands, have considerably larger SD with increasing viewpoint angle. Finally, we can observe that certain joints have consistently large offsets from the baseline skeleton, such as ROOT, HIP\_L, and HIP\_R in Kinect~1 and ANK\_L, ANK\_R, FOO\_L, and FOO\_R, in Kinect~2.

\begin{figure}[!htbp] 
	\begin{minipage}{\columnwidth}
		\begin{tabular}{ccc}
			\includegraphics[trim=0 0 0 0, clip, width=2.7cm]{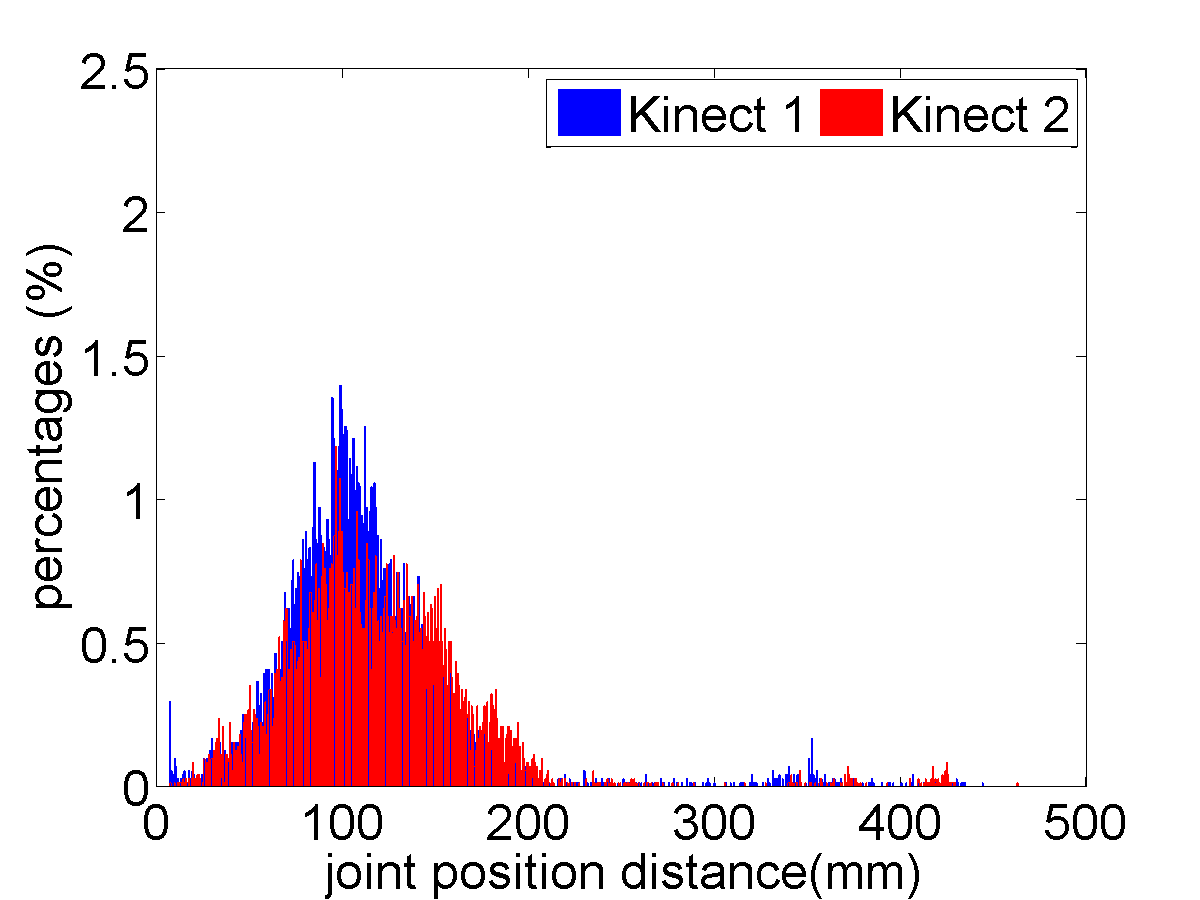}&
			\includegraphics[trim=0 0 0 0, clip, width=2.7cm]{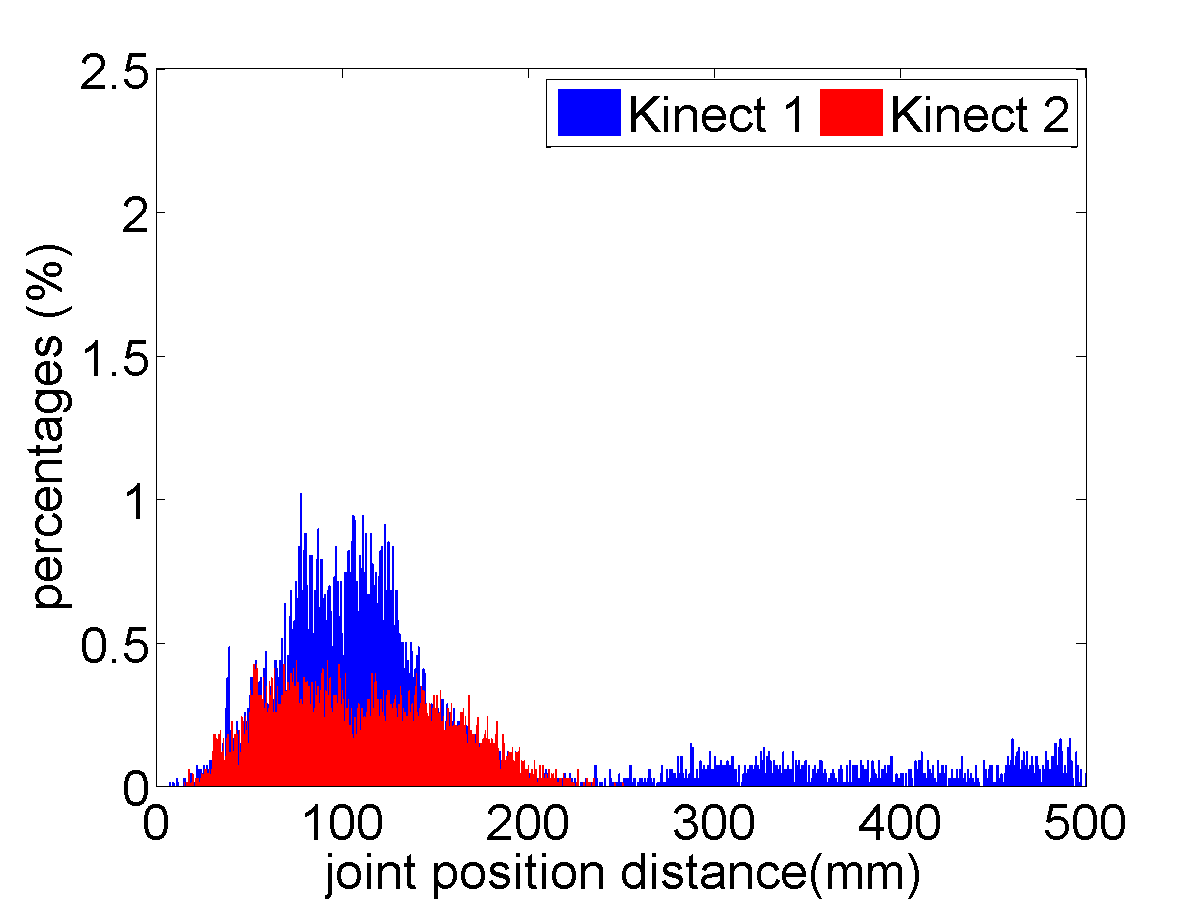}&
			\includegraphics[trim=0 0 0 0, clip, width=2.7cm]{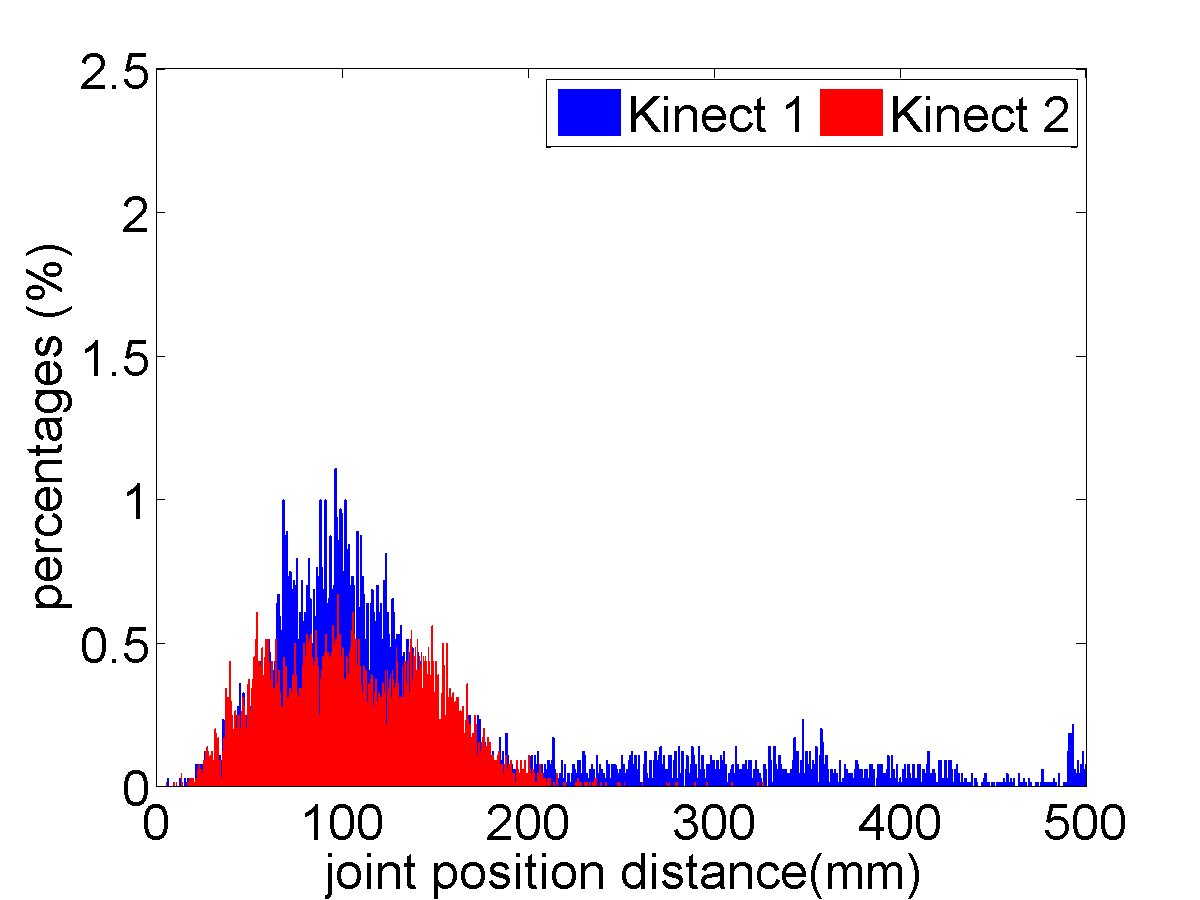}\\
			0$^{\circ}$  & 30$^{\circ}$  & 60$^{\circ}$  \\
		\end{tabular}
		\caption{Distribution of joint position offsets for the right elbow in the exercise \emph{Cops \& Robbers}.}
		\label{fig:joint_pos_err_distribution1}
	\end{minipage}
	\begin{minipage}{\columnwidth}
		\begin{tabular}{ccc}
			\\
			\includegraphics[trim=0 0 0 0, clip, width=2.7cm]{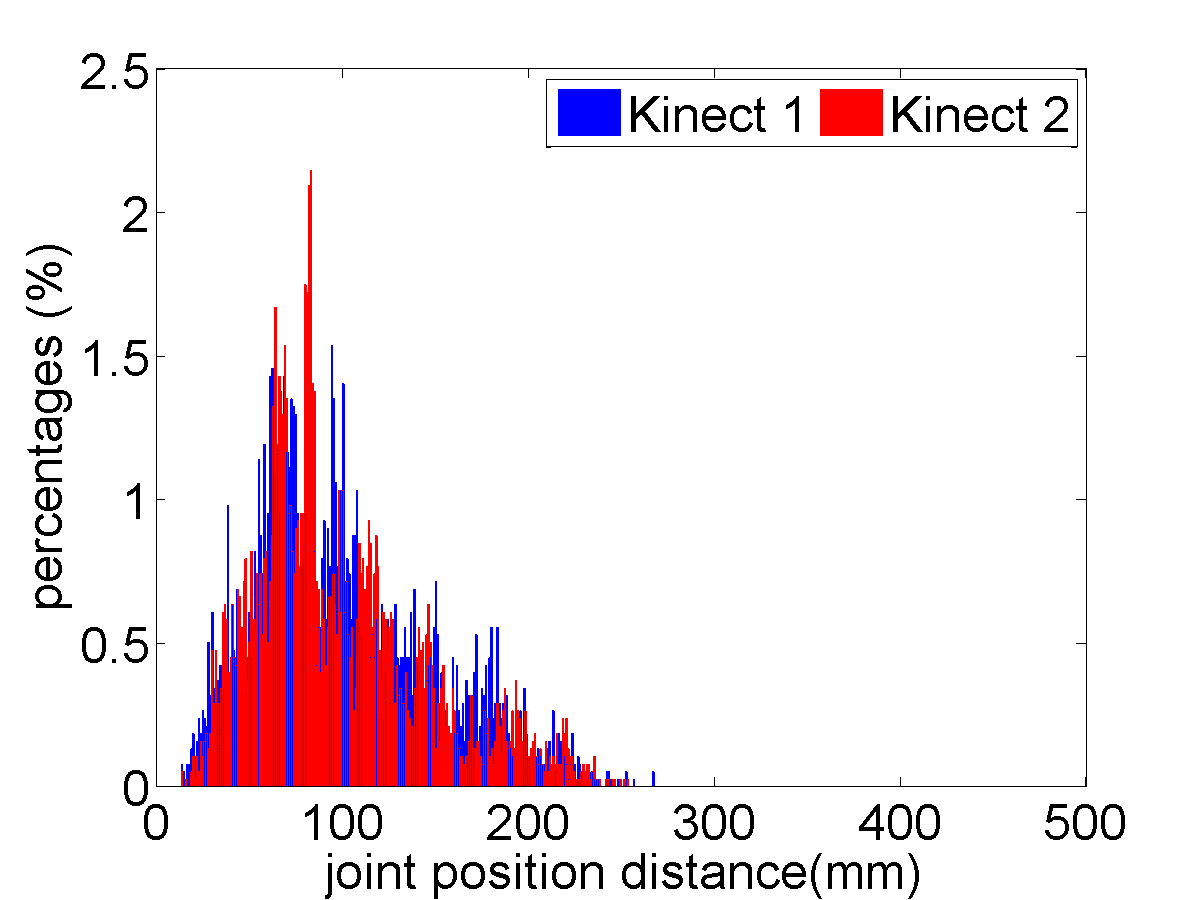}&
			\includegraphics[trim=0 0 0 0, clip, width=2.7cm]{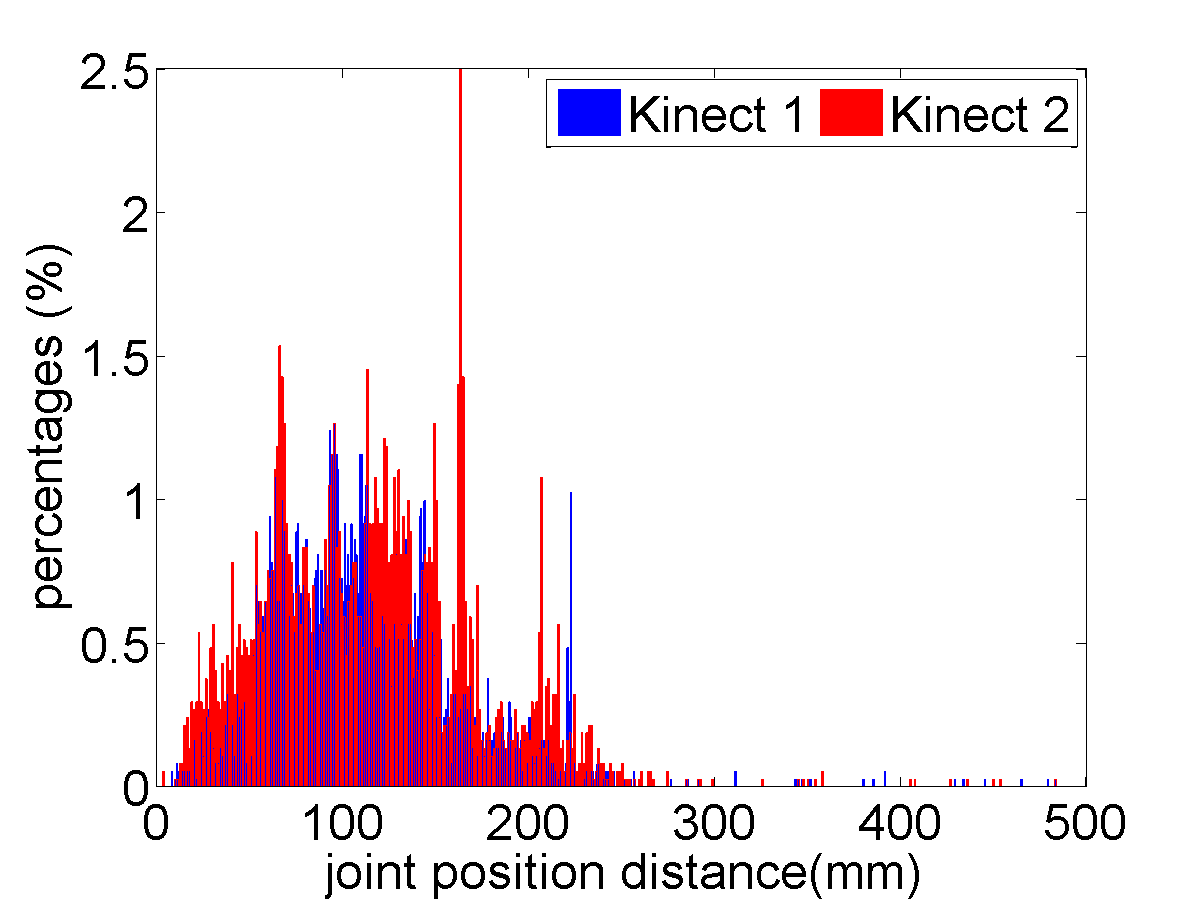}&
			\includegraphics[trim=0 0 0 0, clip, width=2.7cm]{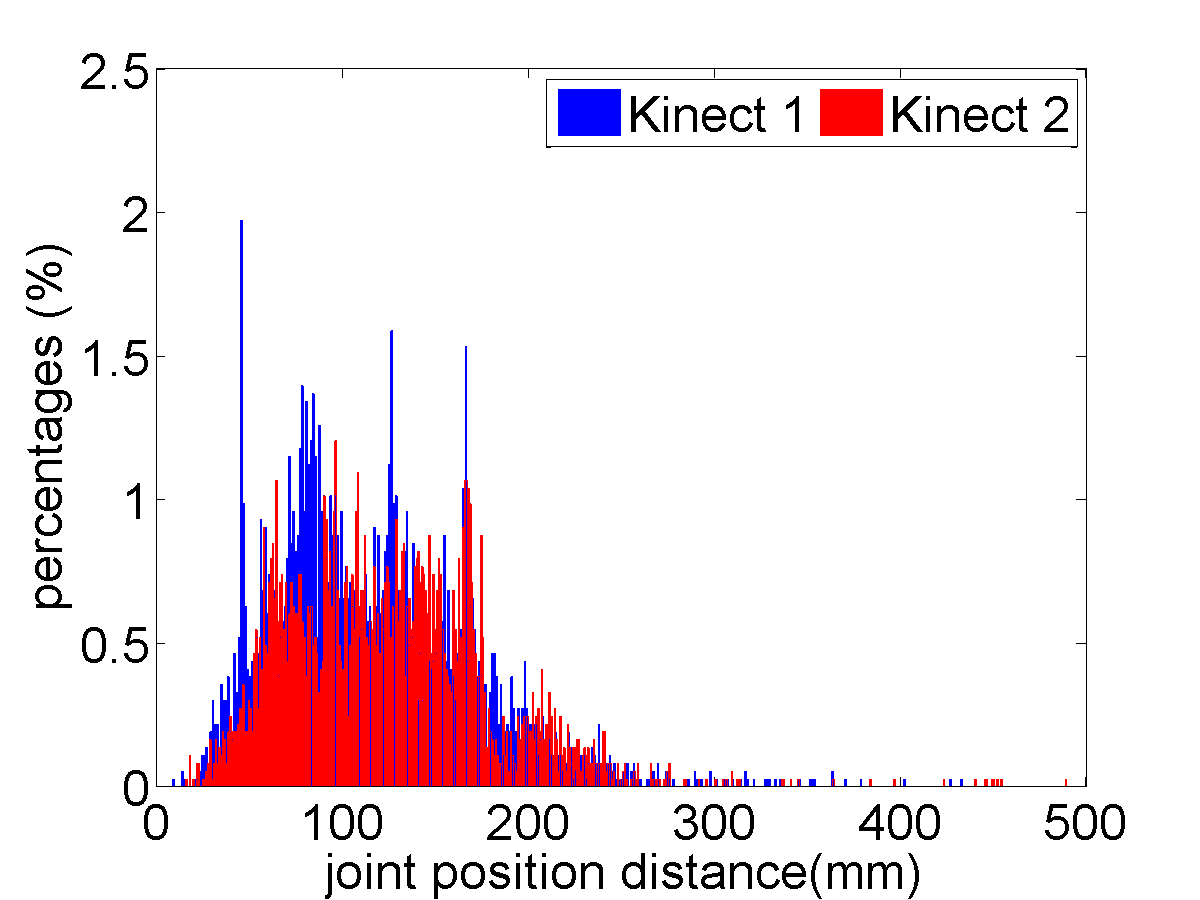}\\
			0$^{\circ}$  & 30$^{\circ}$  & 60$^{\circ}$  \\
		\end{tabular}
		\caption{Distribution of joint position offsets for the right knee in the exercise \emph{Jogging}.}
		\label{fig:joint_pos_err_distribution2}
	\end{minipage}	
\end{figure}

\begin{figure}[!htbp] 
	\begin{tabular}{cc}
		\includegraphics[trim=0 0 0 20, clip, width=4cm]{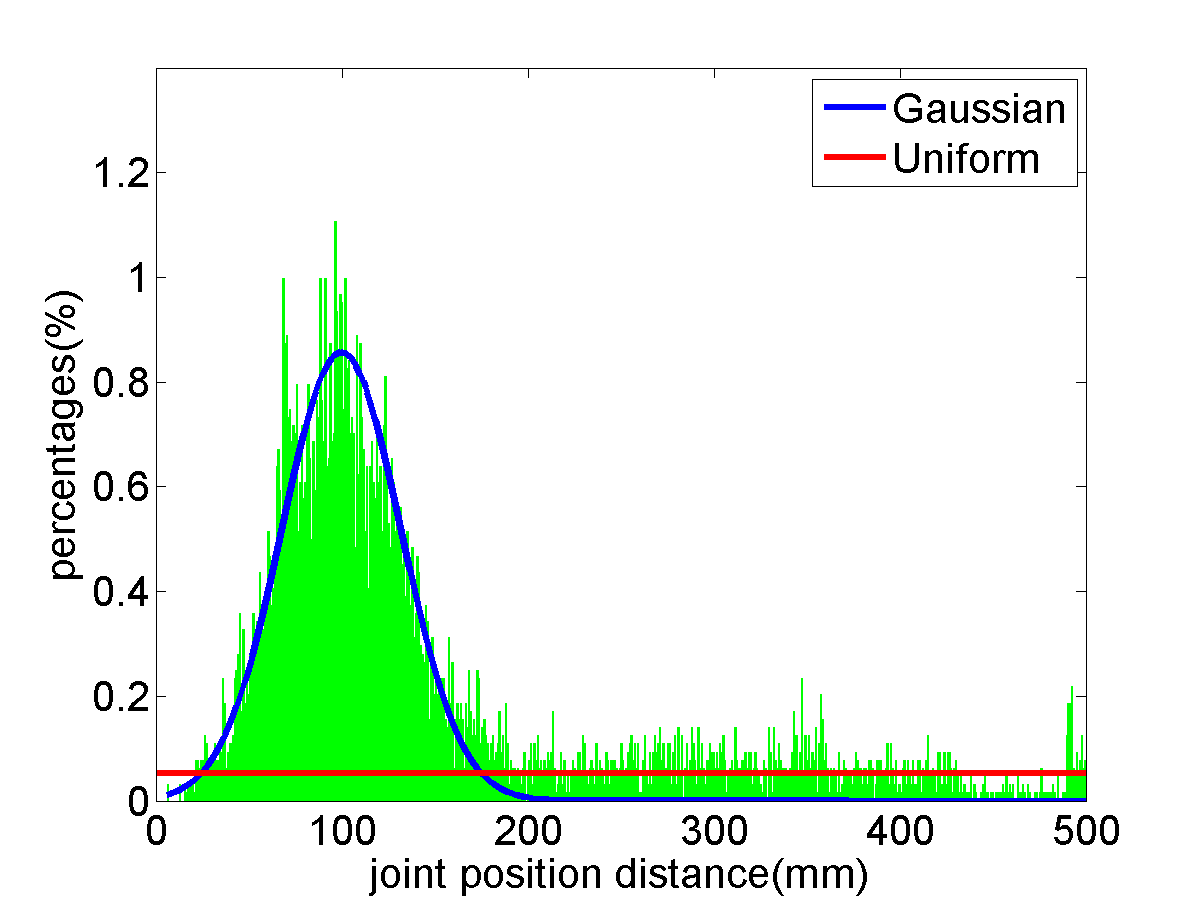}&
		\includegraphics[trim=0 0 0 20, clip, width=4cm]{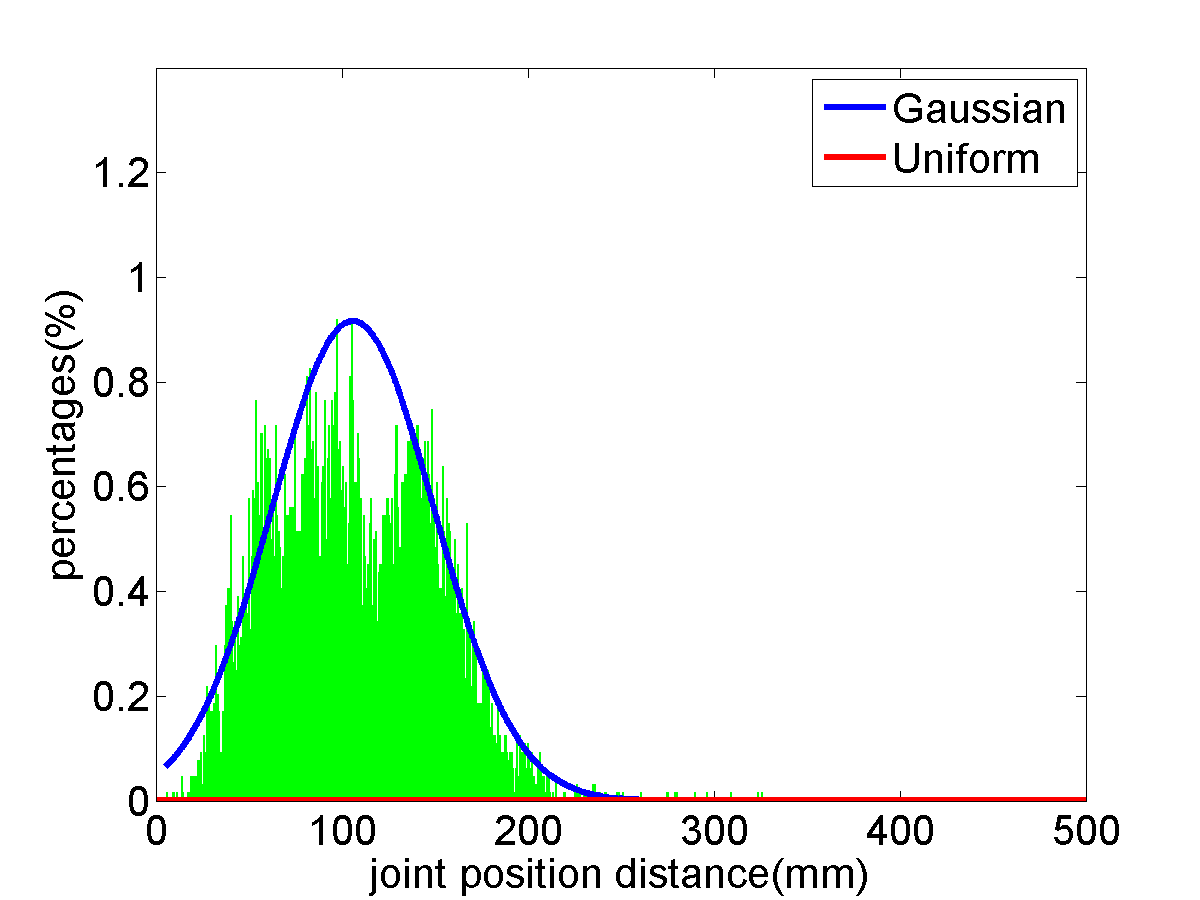}\\
		Kinect~1  & Kinect~2   \\
	\end{tabular}
	\caption{Mixture model fitting into the distribution of joint position offsets for the right elbow in the exercise \emph{Cops and Robbers} (viewpoint: 60$^{\circ}$).}
	\label{fig:joint_pos_err_distribution_fitting}
\end{figure}

The joint position offsets in general depend on various sources of error, such as systematic errors (e.g. offset of hips in Kinect~1), noise from depth computation, occlusions, loss of tracking, etc. In our further analysis, we analyze the error distribution to discriminate between the random errors and the errors due to tracking loss. We expect that the random errors follow Gaussian distribution while the errors due to tracking loss can be treated as outliers belonging to a uniform distribution.
As an example, we show the histogram of the joint position offsets for the right elbow and right knee as captured in the exercises \emph{Cops \& Robbers} and \emph{Jogging} from different viewpoint angles (Figs. \ref{fig:joint_pos_err_distribution1} and \ref{fig:joint_pos_err_distribution2}, respectively). These two joints are among the more active joints in these two exercises. The histograms demonstrate our assumption about the error distribution where the joint position offsets are mainly concentrated close to zero with a long tail to the right side. In order to determine the outliers in the tracking data, we use a mixture model of a Gaussian distribution and a uniform distribution to approximate the distribution of the joint position offsets, as defined in equation~(\ref{equ:mixture_model}). 
Fig.~\ref{fig:joint_pos_err_distribution_fitting} demonstrates the distribution fitting results for the right elbow in the exercise \emph{Cops \& Robbers}. The results show the mixture model of the Gaussian and uniform distributions overlaid on the data histograms.


After applying the mixture model fitting for each joint independently, we can classify the data into either on-track state or off-track state. Table~\ref{tab:joint_pos_outlier_ratio} shows the percentage of the average on-track ratio for each joint defined as the ratio between the number of on-track samples and the total number of frames. The results show that in most joints the on-track ratio is above 90\%. 
For the frontal view, the on-track ratio of all the joint is relatively similar. In the viewpoints of 30$^{\circ}$ and 60$^{\circ}$, the active joints which are further away from the camera typically have lower ratios than the joints closer to the camera. If all the joints in a frame are on-track, that frame is marked as a valid frame for the data accuracy evaluation. The last row in Table~\ref{tab:joint_pos_outlier_ratio} summarized the percentage of valid frames. The percentage of valid frames is typically higher for Kinect 2 than Kinect 1. Furthermore, the percentages of valid frames in the viewpoints of 30$^{\circ}$ and 60$^{\circ}$ drop by 10\% and 15\% compared to those in the frontal view, respectively.
Finally, in Tables~\ref{tab:joint_pos_sitting_outlier} and \ref{tab:joint_pos_standing_outlier} we show the mean and SD of the joint position offsets after the removal of outliers.

\begin{table}[t]
	\centering
	\caption{On-track ratios for Kinect 1 and Kinect 2.}
	\label{tab:joint_pos_outlier_ratio}%
	\begin{tabular}{|l|p{0.2cm}|p{0.2cm}|p{0.2cm}|p{0.2cm}|p{0.2cm}|p{0.2cm}||p{0.2cm}|p{0.2cm}|p{0.2cm}|p{0.2cm}|p{0.2cm}|p{0.2cm}|}
		\hline
		\multirow{3}{*}{} & \multicolumn{6}{c||}{Sitting (\%)}                  & \multicolumn{6}{c|}{Standing (\%)} \\
		\cline{2-13}
		& \multicolumn{3}{c|}{Kinect~1} & \multicolumn{3}{c||}{Kinect~2} & \multicolumn{3}{c|}{Kinect~1} & \multicolumn{3}{c|}{Kinect~2} \\
		\cline{2-13}
		& 0$^{\circ}$     & 30$^{\circ}$    & 60$^{\circ}$    & 0$^{\circ}$     & 30$^{\circ}$    & 60$^{\circ}$    & 0$^{\circ}$     & 30$^{\circ}$    & 60$^{\circ}$    & 0$^{\circ}$     & 30$^{\circ}$    & 60$^{\circ}$ \\
		\hline
		ROOT  & 95    & 97    & 96    & 99    & 98    & 100   & 93    & 97    & 99    & 96    & 99    & 99 \\
		SPINE & 96    & 98    & 98    & 99    & 99    & 100   & 95    & 100   & 99    & 97    & 99    & 99 \\
		NECK  & 96    & 98    & 99    & 99    & 99    & 99    & 98    & 98    & 98    & 97    & 99    & 99 \\
		HEAD  & 96    & 99    & 99    & 100   & 99    & 99    & 97    & 98    & 99    & 97    & 99    & 99 \\
		SHO\_L & 96    & 98    & 99    & 99    & 97    & 95    & 97    & 96    & 95    & 97    & 97    & 97 \\
		ELB\_L & 98    & 98    & 96    & 99    & 97    & 97    & 96    & 94    & 92    & 97    & 96    & 92 \\
		WRI\_L & 97    & 93    & 90    & 99    & 91    & 96    & 97    & 95    & 91    & 97    & 95    & 96 \\
		HAN\_L & 97    & 93    & 92    & 97    & 89    & 93    & 96    & 93    & 89    & 97    & 96    & 95 \\
		SHO\_R & 97    & 97    & 99    & 99    & 98    & 99    & 98    & 98    & 97    & 98    & 99    & 97 \\
		ELB\_R & 98    & 94    & 97    & 99    & 98    & 99    & 97    & 95    & 98    & 98    & 99    & 99 \\
		WRI\_R & 98    & 93    & 91    & 99    & 99    & 99    & 99    & 99    & 99    & 98    & 99    & 99 \\
		HAN\_R & 97    & 92    & 88    & 99    & 98    & 99    & 98    & 99    & 98    & 98    & 99    & 100 \\
		HIP\_L & 96    & 97    & 99    & 99    & 98    & 99    & 95    & 99    & 98    & 95    & 98    & 99 \\
		KNE\_L & 96    & 99    & 96    & 98    & 98    & 97    & 97    & 95    & 87    & 97    & 96    & 91 \\
		ANK\_L & 99    & 98    & 97    & 98    & 98    & 94    & 94    & 93    & 90    & 96    & 91    & 96 \\
		FOO\_L & 99    & 98    & 98    & 97    & 94    & 96    & 95    & 89    & 91    & 96    & 94    & 96 \\
		HIP\_R & 94    & 93    & 98    & 99    & 96    & 100   & 94    & 98    & 99    & 98    & 99    & 100 \\
		KNE\_R & 97    & 96    & 97    & 98    & 97    & 94    & 97    & 98    & 96    & 97    & 98    & 98 \\
		ANK\_R & 99    & 96    & 99    & 97    & 99    & 98    & 94    & 95    & 95    & 97    & 96    & 98 \\
		FOO\_R & 99    & 99    & 98    & 96    & 96    & 95    & 96    & 97    & 95    & 97    & 97    & 99 \\
		\hline
		Overall & 92    & 81    & 75    & 94    & 84    & 81    & 90    & 79    & 73    & 92    & 82    & 80 \\
		\hline
	\end{tabular}%
\end{table}

Compared with the results in Tables~\ref{tab:joint_pos_sitting} and \ref{tab:joint_pos_standing}, both the mean and SD of most joints in Tables~\ref{tab:joint_pos_sitting_outlier} and \ref{tab:joint_pos_standing_outlier} are reduced since the outliers are excluded from the analysis. Table~\ref{tab:joint_pos_outlier_gain} summarized the average reduction of the mean and SD of the joint position offsets after excluding the outliers. The results demonstrate that the data accuracy can be significantly improved by fitting the data into the mixture model.

\begin{table*}[t!]
	\centering
	\hspace{-7mm}
	\begin{minipage}[t]{0.45\linewidth}
		\centering
		\caption{Joint position offsets in sitting exercises with excluded outliers. }
		\label{tab:joint_pos_sitting_outlier}%
		\begin{tabular}{|l|p{0.2cm}|p{0.2cm}|p{0.2cm}|p{0.2cm}|p{0.2cm}|p{0.2cm}||p{0.2cm}|p{0.2cm}|p{0.2cm}|p{0.2cm}|p{0.2cm}|p{0.2cm}|}
			\hline
			\multirow{3}{*}{} & \multicolumn{6}{c||}{Kinect~1}                  & \multicolumn{6}{c|}{Kinect~2} \\
			\cline{2-13}
			& \multicolumn{3}{c|}{Mean (mm)} & \multicolumn{3}{c||}{SD (mm)} & \multicolumn{3}{c|}{Mean (mm)} & \multicolumn{3}{c|}{SD (mm)} \\
			\cline{2-13}
			& 0$^{\circ}$     & 30$^{\circ}$    & 60$^{\circ}$    & 0$^{\circ}$     & 30$^{\circ}$    & 60$^{\circ}$    & 0$^{\circ}$     & 30$^{\circ}$    & 60$^{\circ}$    & 0$^{\circ}$     & 30$^{\circ}$    & 60$^{\circ}$ \\
			\hline
			ROOT$**$  \footnotetext[0]{* \textit{t}-test, p $<$ 0.05, ** \textit{t}-test, p $<$ 0.01} & 251   & 259   & 261   & 15    & 13    & 17    & 100   & 100   & 107   & 14    & 12    & 14  \\
			SPINE & 85    & 92    & 96    & 13    & 12    & 14    & 110   & 116   & 128   & 11    & 10    & 11  \\
			NECK  & 73    & 61    & 60    & 14    & 14    & 16    & 84    & 73    & 71    & 12    & 11    & 12  \\
			HEAD  & 69    & 66    & 63    & 15    & 14    & 15    & 49    & 47    & 50    & 11    & 10    & 11  \\
			SHO\_L & 84    & 83    & 92    & 14    & 16    & 24    & 74    & 71    & 78    & 13    & 16    & 23  \\
			ELB\_L & 77    & 80    & 92    & 18    & 17    & 22    & 86    & 95    & 109   & 13    & 14    & 19  \\
			WRI\_L & 70    & 75    & 111   & 20    & 24    & 29    & 55    & 70    & 109   & 15    & 19    & 26  \\
			HAN\_L & 78    & 86    & 123   & 28    & 30    & 38    & 60    & 79    & 117   & 19    & 20    & 28  \\
			SHO\_R & 71    & 70    & 68    & 16    & 15    & 16    & 79    & 82    & 77    & 14    & 13    & 15  \\
			ELB\_R & 90    & 80    & 83    & 18    & 18    & 22    & 88    & 77    & 71    & 17    & 17    & 17  \\
			WRI\_R & 59    & 70    & 89    & 19    & 22    & 30    & 59    & 60    & 71    & 16    & 16    & 18  \\
			HAN\_R & 79    & 88    & 105   & 23    & 27    & 35    & 72    & 66    & 75    & 14    & 16    & 19  \\
			HIP\_L$**$ & 182   & 197   & 214   & 15    & 13    & 15    & 115   & 120   & 138   & 13    & 12    & 14  \\
			KNE\_L & 92    & 90    & 86    & 17    & 18    & 19    & 76    & 88    & 104   & 14    & 14    & 20  \\
			ANK\_L$*$ & 52    & 72    & 75    & 13    & 17    & 21    & 93    & 99    & 107   & 13    & 13    & 18  \\
			FOO\_L$*$ & 66    & 70    & 81    & 17    & 19    & 27    & 91    & 100   & 110   & 14    & 17    & 27  \\
			HIP\_R$**$ & 200   & 206   & 210   & 14    & 13    & 16    & 132   & 127   & 133   & 13    & 12    & 13  \\
			KNE\_R & 120   & 113   & 123   & 16    & 16    & 25    & 120   & 114   & 111   & 12    & 12    & 16  \\
			ANK\_R$*$ & 66    & 72    & 88    & 13    & 14    & 20    & 112   & 121   & 127   & 13    & 15    & 18  \\
			FOO\_R$*$ & 66    & 72    & 81    & 13    & 16    & 18    & 125   & 134   & 133   & 14    & 15    & 19  \\
			\hline
		\end{tabular}%
	\end{minipage}
	\hspace{10mm}
	\begin{minipage}[t]{0.45\linewidth}
		\centering
		\caption{Joint position offsets in standing exercises with excluded outliers. }
		\label{tab:joint_pos_standing_outlier}%
		\begin{tabular}{|l|p{0.2cm}|p{0.2cm}|p{0.2cm}|p{0.2cm}|p{0.2cm}|p{0.2cm}||p{0.2cm}|p{0.2cm}|p{0.2cm}|p{0.2cm}|p{0.2cm}|p{0.2cm}|}
			\hline
			\multirow{3}{*}{} & \multicolumn{6}{c||}{Kinect~1}                  & \multicolumn{6}{c|}{Kinect~2} \\
			\cline{2-13}
			& \multicolumn{3}{c|}{Mean (mm)} & \multicolumn{3}{c||}{SD (mm)} & \multicolumn{3}{c|}{Mean (mm)} & \multicolumn{3}{c|}{SD (mm)} \\
			\cline{2-13}
			& 0$^{\circ}$     & 30$^{\circ}$    & 60$^{\circ}$    & 0$^{\circ}$     & 30$^{\circ}$    & 60$^{\circ}$    & 0$^{\circ}$     & 30$^{\circ}$    & 60$^{\circ}$    & 0$^{\circ}$     & 30$^{\circ}$    & 60$^{\circ}$ \\
			\hline
			ROOT$**$ \footnotetext[0]{* \textit{t}-test, p $<$ 0.05, ** \textit{t}-test, p $<$ 0.01} & 244   & 257   & 267   & 16    & 17    & 20    & 67    & 80    & 91    & 10    & 11    & 12  \\
			SPINE & 76    & 87    & 99    & 14    & 16    & 17    & 108   & 130   & 143   & 9     & 10    & 10  \\
			NECK  & 81    & 80    & 84    & 18    & 20    & 19    & 73    & 79    & 88    & 13    & 14    & 15  \\
			HEAD  & 86    & 87    & 87    & 17    & 19    & 18    & 58    & 73    & 77    & 11    & 12    & 13  \\
			SHO\_L & 68    & 67    & 70    & 20    & 23    & 26    & 60    & 64    & 71    & 15    & 18    & 20  \\
			ELB\_L & 88    & 100   & 109   & 26    & 32    & 35    & 101   & 125   & 136   & 17    & 18    & 23  \\
			WRI\_L & 70    & 77    & 85    & 26    & 29    & 34    & 50    & 66    & 80    & 17    & 20    & 24  \\
			HAN\_L & 68    & 78    & 84    & 26    & 31    & 35    & 62    & 78    & 89    & 16    & 17    & 22  \\
			SHO\_R & 70    & 72    & 70    & 20    & 23    & 21    & 93    & 107   & 109   & 16    & 16    & 16  \\
			ELB\_R & 92    & 83    & 76    & 30    & 31    & 28    & 90    & 86    & 78    & 24    & 22    & 23  \\
			WRI\_R & 73    & 74    & 74    & 31    & 29    & 27    & 72    & 85    & 86    & 21    & 18    & 19  \\
			HAN\_R & 77    & 74    & 70    & 32    & 33    & 31    & 73    & 73    & 71    & 18    & 18    & 18  \\
			HIP\_L$**$ & 183   & 203   & 224   & 17    & 19    & 19    & 100   & 124   & 145   & 11    & 11    & 13  \\
			KNE\_L & 95    & 103   & 108   & 18    & 23    & 23    & 103   & 122   & 132   & 14    & 15    & 18  \\
			ANK\_L$*$ & 110   & 130   & 119   & 23    & 28    & 25    & 133   & 157   & 160   & 17    & 19    & 21  \\
			FOO\_L$*$ & 106   & 120   & 105   & 24    & 31    & 26    & 126   & 156   & 165   & 18    & 23    & 25  \\
			HIP\_R$**$ & 184   & 196   & 207   & 16    & 17    & 19    & 95    & 102   & 109   & 10    & 10    & 12  \\
			KNE\_R & 99    & 98    & 98    & 18    & 19    & 19    & 109   & 121   & 118   & 15    & 17    & 17  \\
			ANK\_R$*$ & 106   & 107   & 108   & 23    & 23    & 22    & 144   & 161   & 156   & 19    & 20    & 20  \\
			FOO\_R$*$ & 92    & 85    & 85    & 23    & 23    & 24    & 142   & 157   & 151   & 20    & 20    & 24  \\
			\hline
		\end{tabular}%
	\end{minipage}
\end{table*}

\begin{table}[!htbp] 
	\centering
	\caption{Average joint position offsets without outliers. }
	\begin{tabular}{|l|p{0.2cm}|p{0.2cm}|p{0.2cm}|p{0.2cm}|p{0.2cm}|p{0.2cm}||p{0.2cm}|p{0.2cm}|p{0.2cm}|p{0.2cm}|p{0.2cm}|p{0.2cm}|}
		\hline
		\multirow{3}{*}{} & \multicolumn{6}{c||}{Kinect~1}                  & \multicolumn{6}{c|}{Kinect~2} \\
		\cline{2-13}
		& \multicolumn{3}{c|}{Mean (\%)} & \multicolumn{3}{c||}{SD (\%)} & \multicolumn{3}{c|}{Mean (\%)} & \multicolumn{3}{c|}{SD (\%)} \\
		\cline{2-13}
		& 0$^{\circ}$     & 30$^{\circ}$    & 60$^{\circ}$    & 0$^{\circ}$     & 30$^{\circ}$    & 60$^{\circ}$    & 0$^{\circ}$     & 30$^{\circ}$    & 60$^{\circ}$    & 0$^{\circ}$     & 30$^{\circ}$    & 60$^{\circ}$ \\
		\hline
		Sitting & 5     & 8     & 6     & 35    & 38    & 35    & 2     & 5     & 3     & 23    & 36    & 23  \\
		Standing & 8     & 7     & 8     & 39    & 36    & 33    & 8     & 5     & 6     & 49    & 46    & 43  \\
		\hline
	\end{tabular}%
	\label{tab:joint_pos_outlier_gain}%
\end{table}%

\subsection{Bone Length Accuracy}

Another important parameter for evaluation of Kinect pose tracking performance is the bone length. As mentioned previously, the Kinect tracking algorithm does not specifically pre-define or calibrate for the anthropometric values of the body/bone segments. On the other hand, the human skeleton can be approximated as a kinematic structure with rigid segments, so we expect that the bone lengths should stay relatively constant. The size of the variance (and SD) of the bone length over time can thus be interpreted as a measure of robustness of the extracted kinematic model.

\begin{figure*}[!htbp] 
	\begin{tabular}{ccc}
		\includegraphics[width=5.5cm]{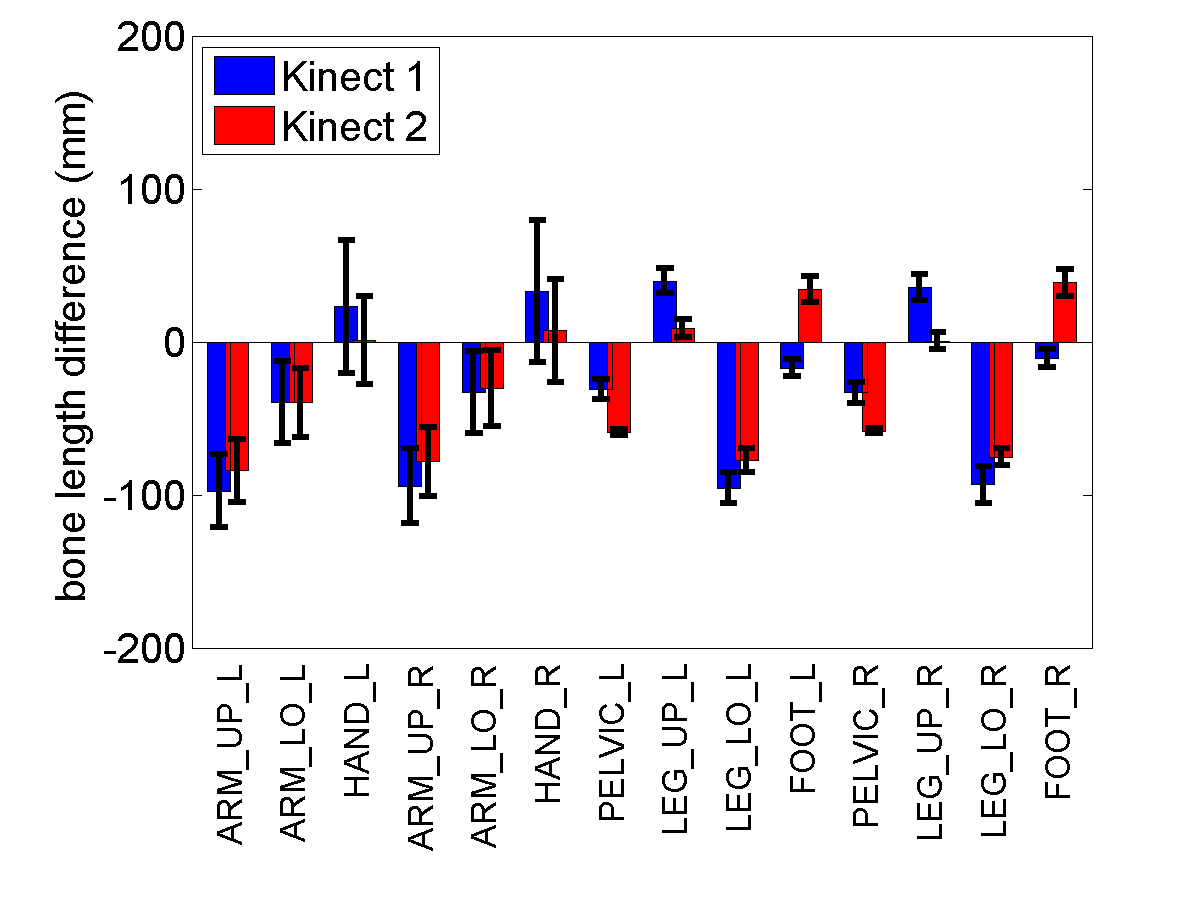}&
		\includegraphics[width=5.5cm]{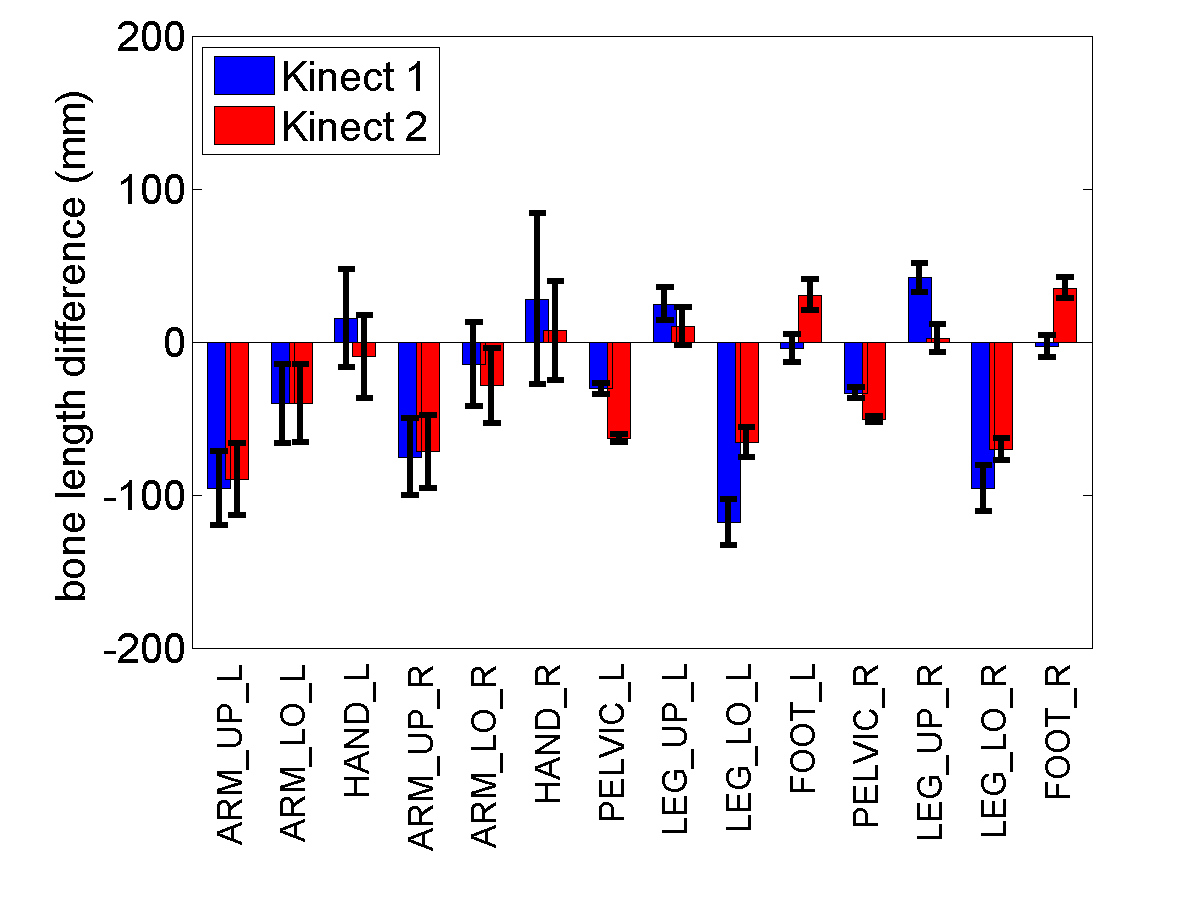}&
		\includegraphics[width=5.5cm]{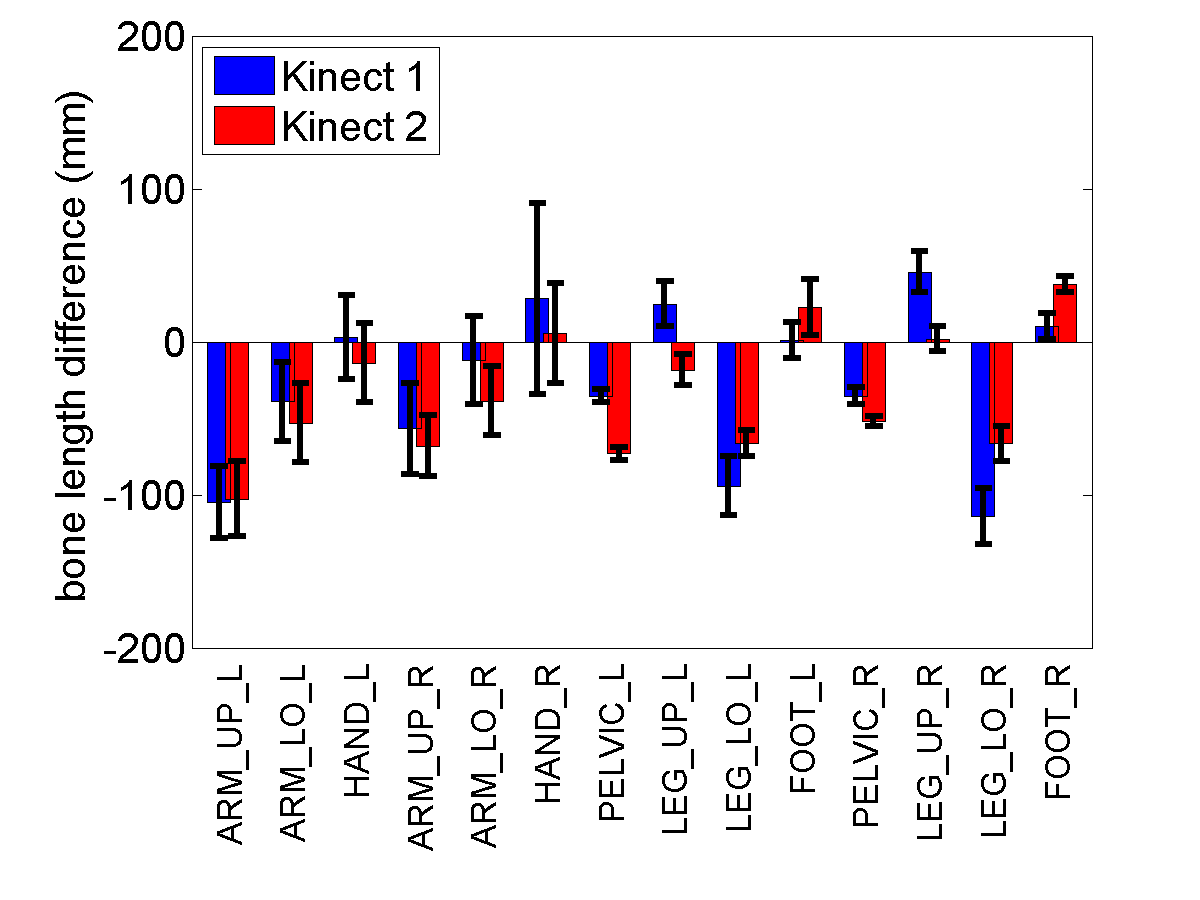}\\
		0$^{\circ}$ &30$^{\circ}$ &60$^{\circ}$
	\end{tabular}
	\caption{Mean bone length differences and the corresponding SDs for the exercise \emph{Cops \& Robbers} as captured by Kinect~1 and Kinect~2 from three different viewpoints.}
	\label{fig:bone_length_mean_sd_1}
\end{figure*}

\begin{figure*}[!htbp] 
	\begin{tabular}{ccc}
		\includegraphics[width=5.5cm]{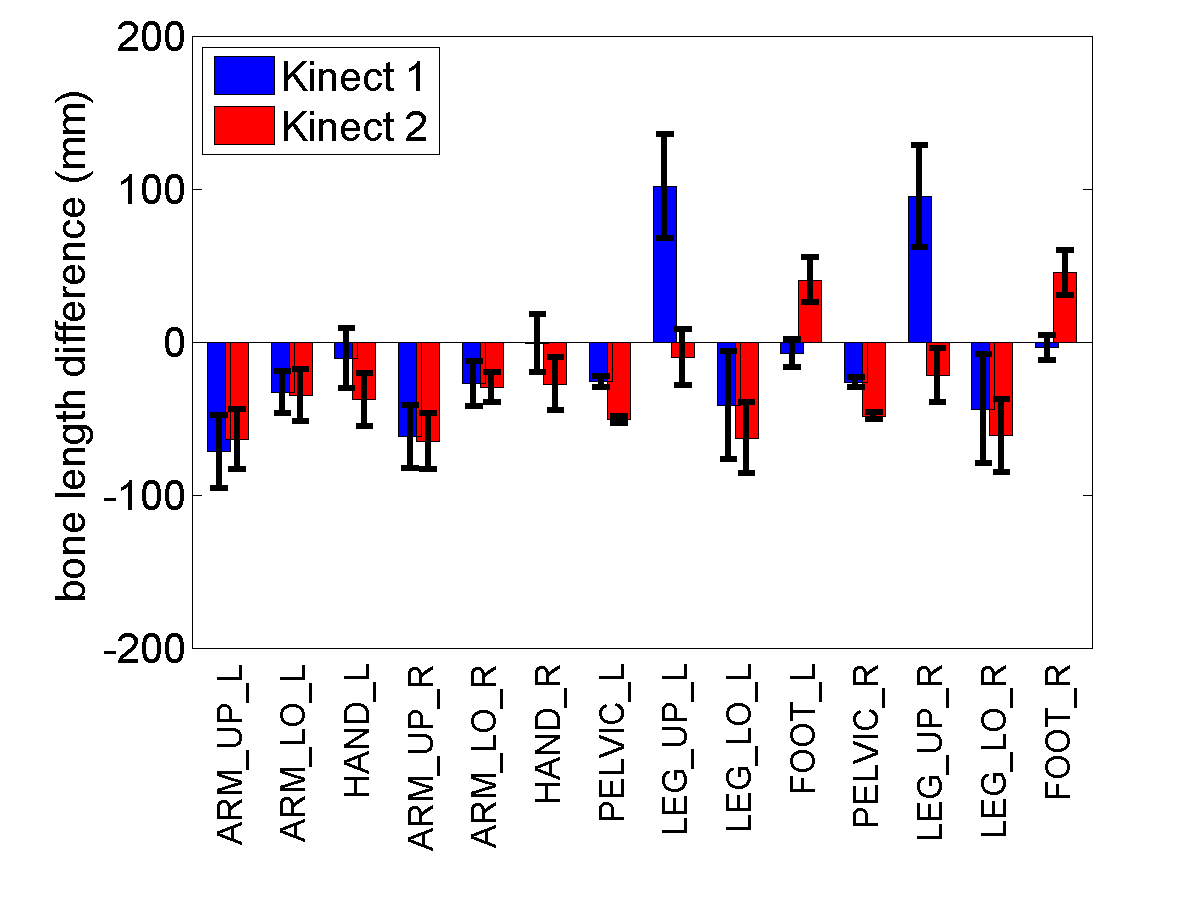}&
		\includegraphics[width=5.5cm]{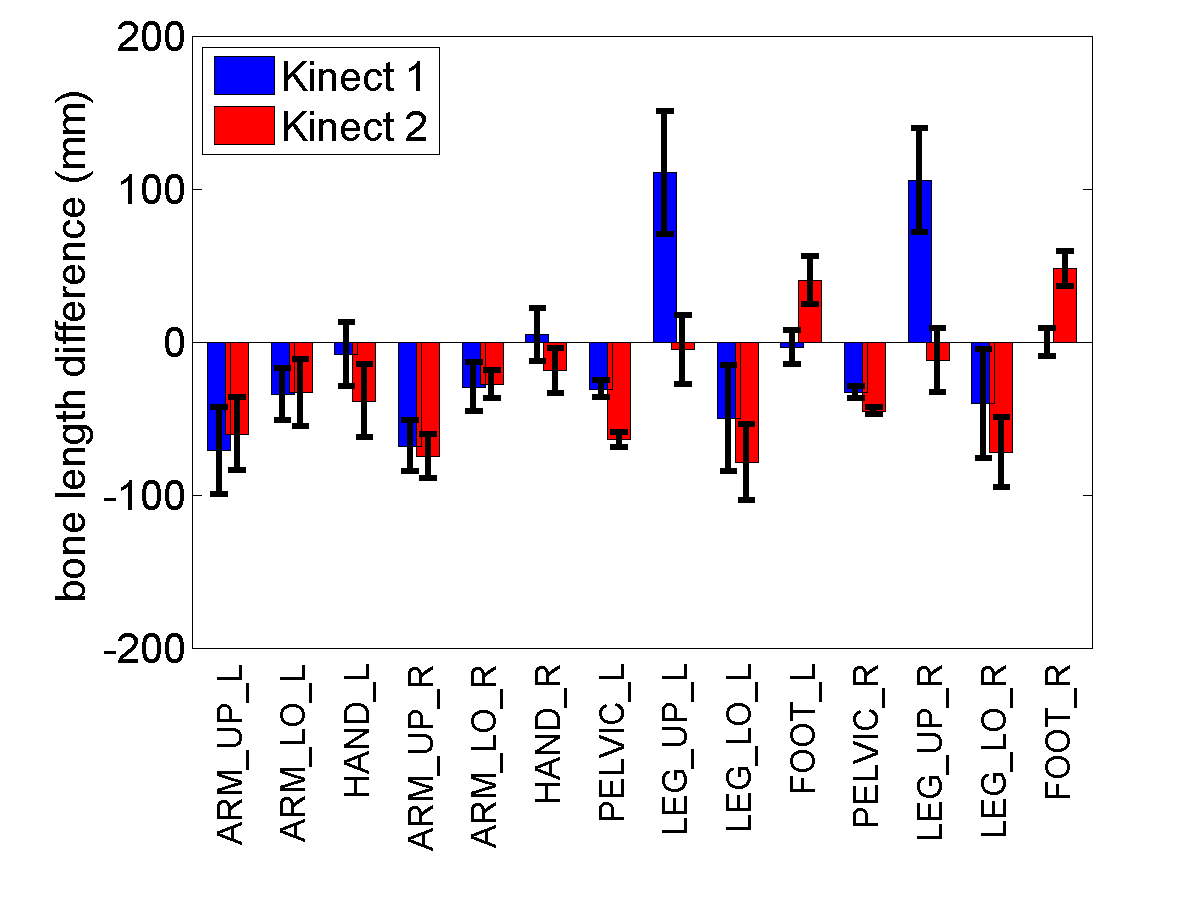}&
		\includegraphics[width=5.5cm]{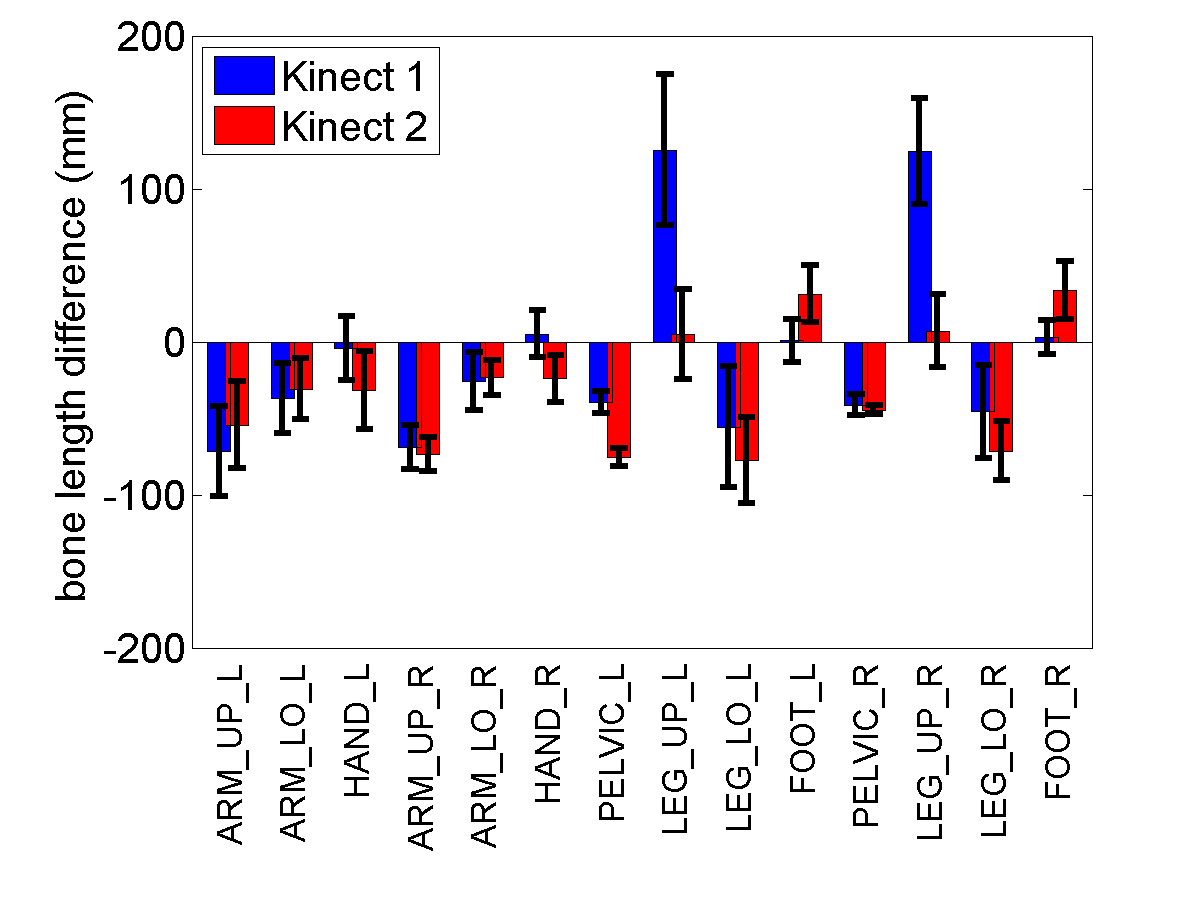}\\
		0$^{\circ}$ &30$^{\circ}$ &60$^{\circ}$
	\end{tabular}
	\caption{Mean bone length differences and the corresponding SDs for the exercise \emph{Jogging} as captured by Kinect~1 and Kinect~2 from three different viewpoints.}
	\label{fig:bone_length_mean_sd_2}
\end{figure*}

For the Kinect skeleton, we define the bone length as the $l_2$ distance between the positions of two subsequent joints. The bone length for the motion capture is on the other hand determined during the calibration phase and remains constant during the motion sequence. Figs. \ref{fig:bone_length_mean_sd_1} and \ref{fig:bone_length_mean_sd_2} show the means and SDs of the bone length difference of Kinect~1 and Kinect~2, respectively, as compared to the bone length calibrated from the motion capture data across all the subjects.
The mean bone length difference does not change too much between different exercises. The SDs are typically increasing with larger viewpoint angle.
We can observe that the bone lengths in Kinect~1 usually have larger offsets and SDs as compared to Kinect~2, especially for the upper legs due to the large vertical offset of the hip joints. 

Tables \ref{tab:bone_length_sitting} and \ref{tab:bone_length_standing} summarize the mean and SD of the bone length differences in Kinect~1 and Kinect~2 in the three viewpoints for sitting and standing exercises, receptively. We can observe that the mean differences in the bone lengths and SDs are smaller in Kinect~2, suggesting that the kinematic structure of its skeleton is more robust.

\begin{table*}[!htbp] 
	\begin{minipage}{1.0\linewidth}
		\centering
		\caption{Mean and SD of the bone length difference, sitting pose}
		\label{tab:bone_length_sitting}%
		\begin{tabular}{|l|c|c|c|c|c|c||c|c|c|c|c|c|}
			\hline
			\multirow{3}{*}{} & \multicolumn{6}{c||}{Kinect~1}                  & \multicolumn{6}{c|}{Kinect~2} \\
			\cline{2-13}
			& \multicolumn{3}{c|}{Mean (mm)} & \multicolumn{3}{c||}{SD (mm)} & \multicolumn{3}{c|}{Mean (mm)} & \multicolumn{3}{c|}{SD (mm)} \\
			\cline{2-13}
			& 0$^{\circ}$     & 30$^{\circ}$    & 60$^{\circ}$    & 0$^{\circ}$     & 30$^{\circ}$    & 60$^{\circ}$    & 0$^{\circ}$     & 30$^{\circ}$    & 60$^{\circ}$    & 0$^{\circ}$     & 30$^{\circ}$    & 60$^{\circ}$ \\
			\hline
			ARM\_UP\_L \footnotetext[0]{\hspace{32mm} * \textit{t}-test, p $<$ 0.05, ** \textit{t}-test, p $<$ 0.01} & -76   & -74   & -81   & 17    & 18    & 21    & -67   & -62   & -68   & 14    & 17    & 20 \\
			ARM\_LO\_L & -46   & -40   & -32   & 17    & 22    & 27    & -39   & -30   & -24   & 14    & 19    & 23 \\
			HAND\_L$**$ & -3    & -4    & -3    & 22    & 19    & 20    & -20   & -17   & -17   & 20    & 24    & 25 \\
			ARM\_UP\_R & -68   & -70   & -69   & 17    & 17    & 18    & -67   & -67   & -69   & 15    & 14    & 13 \\
			ARM\_LO\_R & -37   & -30   & -29   & 17    & 20    & 21    & -35   & -28   & -30   & 13    & 15    & 16 \\
			HAND\_R$**$ & 2     & 3     & 3     & 21    & 24    & 24    & -15   & -13   & -9    & 20    & 19    & 17 \\
			PELVIC\_L$**$ & -28   & -28   & -36   & 6     & 5     & 8     & -54   & -62   & -75   & 3     & 3     & 4 \\
			LEG\_UP\_L & 49    & 38    & 38    & 30    & 32    & 34    & 9     & 5     & 5     & 16    & 18    & 18 \\
			LEG\_LO\_L & -90   & -100  & -81   & 23    & 28    & 31    & -80   & -79   & -75   & 17    & 18    & 21 \\
			FOOT\_L$**$ & -13   & -4    & 1     & 8     & 10    & 13    & 36    & 38    & 32    & 12    & 14    & 19 \\
			PELVIC\_R$*$ & -29   & -33   & -41   & 6     & 5     & 7     & -52   & -47   & -49   & 3     & 3     & 4 \\
			LEG\_UP\_R & 44    & 46    & 53    & 31    & 28    & 31    & 1     & 9     & 18    & 16    & 18    & 18 \\
			LEG\_LO\_R & -86   & -88   & -95   & 25    & 26    & 31    & -76   & -82   & -74   & 17    & 19    & 23 \\
			FOOT\_R$**$ & -8    & -2    & 8     & 7     & 9     & 11    & 41    & 44    & 40    & 12    & 11    & 12 \\
			\hline
		\end{tabular}%
	\end{minipage}
\end{table*}%

\begin{table*}[!htbp] 
	\begin{minipage}{1.0\linewidth}
		\centering
		\caption{Mean and SD of the bone length difference, standing pose}
		\label{tab:bone_length_standing}%
		\begin{tabular}{|l|c|c|c|c|c|c||c|c|c|c|c|c|}
			\hline
			\multirow{3}{*}{} & \multicolumn{6}{c||}{Kinect~1}                  & \multicolumn{6}{c|}{Kinect~2} \\
			\cline{2-13}
			& \multicolumn{3}{c|}{Mean (mm)} & \multicolumn{3}{c||}{SD (mm)} & \multicolumn{3}{c|}{Mean (mm)} & \multicolumn{3}{c|}{SD (mm)} \\
			\cline{2-13}
			& 0$^{\circ}$     & 30$^{\circ}$    & 60$^{\circ}$    & 0$^{\circ}$     & 30$^{\circ}$    & 60$^{\circ}$    & 0$^{\circ}$     & 30$^{\circ}$    & 60$^{\circ}$    & 0$^{\circ}$     & 30$^{\circ}$    & 60$^{\circ}$ \\
			\hline
			ARM\_UP\_L$*$ \footnotetext[0]{\hspace{32mm} * \textit{t}-test, p $<$ 0.05, ** \textit{t}-test, p $<$ 0.01} & -60   & -62   & -65   & 19    & 22    & 23    & -54   & -54   & -51   & 15    & 18    & 20 \\
			ARM\_LO\_L$*$ & -30   & -31   & -34   & 16    & 18    & 20    & -34   & -33   & -32   & 17    & 17    & 18 \\
			HAND\_L$**$ & -7    & -5    & 0     & 21    & 21    & 23    & -32   & -32   & -27   & 20    & 21    & 23 \\
			ARM\_UP\_R & -53   & -58   & -59   & 21    & 19    & 18    & -59   & -66   & -66   & 17    & 14    & 14 \\
			ARM\_LO\_R & -22   & -22   & -22   & 17    & 18    & 20    & -30   & -27   & -25   & 18    & 18    & 17 \\
			HAND\_R$**$ & -1    & 4     & 6     & 23    & 21    & 19    & -21   & -16   & -14   & 19    & 18    & 18 \\
			PELVIC\_L$**$ & -27   & -30   & -37   & 9     & 7     & 8     & -52   & -63   & -75   & 5     & 5     & 5 \\
			LEG\_UP\_L$**$ & 113   & 119   & 137   & 28    & 33    & 39    & -13   & -10   & 4     & 19    & 21    & 25 \\
			LEG\_LO\_L & -64   & -66   & -65   & 29    & 29    & 31    & -79   & -85   & -86   & 21    & 23    & 25 \\
			FOOT\_L$**$ & -6    & -1    & 2     & 9     & 12    & 12    & 43    & 42    & 34    & 9     & 12    & 14 \\
			PELVIC\_R$**$ & -28   & -34   & -41   & 7     & 6     & 8     & -47   & -44   & -44   & 7     & 4     & 5 \\
			LEG\_UP\_R$**$ & 103   & 115   & 128   & 29    & 27    & 29    & -25   & -18   & -6    & 18    & 18    & 20 \\
			LEG\_LO\_R & -61   & -57   & -58   & 26    & 25    & 25    & -79   & -79   & -77   & 20    & 20    & 20 \\
			FOOT\_R$**$ & -2    & 1     & 4     & 8     & 9     & 10    & 48    & 45    & 32    & 9     & 11    & 16 \\
			\hline
		\end{tabular}%
	\end{minipage}
\end{table*}%

\subsection{Summary of Findings}
Based on the experimental results reported in this paper, we can make the following observations:

\begin{itemize}[leftmargin=*]
	
	\item As reported by other researchers, the hip joints in Kinect~1 are located much higher than normal with the offset of about 200 mm. The offsets should be considered when calculating knee and hip angles, in particular in sitting position. On the other hand, the skeleton in Kinect~2 is in general more anthropometric with smaller offsets.
	
	\item The foot and ankle joints of Kinect~2 are offset from the ground plane for about 100 mm or more. The orientation of the feet is thus unreliable. Once the foot is lifted off the ground, the tracking of the joints is more accurate. The unreliable foot position may be originating from ToF artifacts that generate large amounts of noise close to large planar surfaces.
	
	\item Overall accuracy of joint positions in Kinect~2 is better than in Kinect~1, except the location of feet. The average offsets are typically between 50 mm and 100 mm.
	
	\item The analysis of the distribution mixture shows that Kinect~2 has smaller uniform distribution component (i.e. less outliers) suggesting that the tracking of joints is more robust. Kinect~2 also tracks human movement more reliably even with partial body occlusions.
	
	\item The difference and variance of the actual limb lengths are smaller in Kinect~2 than in Kinect~1.
	
	\item The skeleton tracking in Kinect~2 has much smaller latency as compared to Kinect~1 which is noticeable especially during fast actions (e.g. exercises \emph{Jogging} and \emph{Punching}). 
	
\end{itemize}

	\section{Conclusion}
	\label{sec:conclusions}
In this paper, we compared the human pose estimation for the first and second generations of Microsoft Kinect with standard motion capture technology. The results of our analysis showed that overall Kinect~2 has better accuracy in joint estimation while providing skeletal tracking that is more robust to occlusions and body rotation. Only the lower legs were tracked with large offsets, possibly due to ToF artifacts. This phenomena was not observed in Kinect~1 which employs structured light for depth acquisition. For Kinect~1, the largest offsets were observed in the pelvic area as also noted by others. Our analyses show that Kinect~1 can be exchanged with Kinect~2 for majority of motions.
Furthermore, by applying a mixture of Gaussian and uniform distribution models we were able to evaluate the robustness of the pose tracking. We showed that the SDs of the joint positions can be reduced by 30\% to 40\% by employing the classification with a mixture distribution model. This finding suggests that by excluding the outliers from the data and compensating for the offsets, more accurate human motion analysis can be achieved.

\bibliographystyle{IEEEtran}
\bibliography{biblio}

\end{document}